%% file: emnlp2020.tex
\documentclass[11pt,a4paper]{article}
\usepackage[nohyperref]{emnlp2020}
\usepackage{times}
\usepackage{latexsym}

\usepackage{booktabs}
\usepackage{graphicx}

\usepackage{microtype}

\aclfinalcopy 

\newcommand{\conv}{Conv\-AI2}

\title{Controlling Style in Generated Dialogue}

\author{
  \qquad \quad Eric Michael Smith*, Diana Gonzalez-Rico*, Emily Dinan, Y-Lan Boureau\\
  \qquad \quad Facebook AI Research \\
}

\date{}

\begin{document}
\maketitle
\begin{abstract}
Open-domain conversation models have become good at generating natural-sounding dialogue,
using very large architectures with billions of trainable parameters. 
The vast training data required to train these architectures aggregates many different styles, tones, and qualities. Using that data to train 
a single model makes it difficult to use
the model as a consistent conversational agent, e.g. with 
a stable set of persona traits and a typical style of expression. 
Several architectures affording
control mechanisms over generation architectures
have been proposed, each with different trade-offs. However, it remains unclear
whether their use in dialogue is
viable, and what the trade-offs look like
with the most recent state-of-the-art conversational architectures.
In this work, we adapt three 
previously proposed controllable generation
architectures to open-domain dialogue generation, controlling the style
of the generation to match one among
about 200 possible styles. We compare their 
respective performance and trade-offs,
and show how they can be used to provide
 insights into existing conversational
datasets, and generate a varied set of styled conversation
replies.
\end{abstract}

\section{Introduction}

Conversational models have shown
vastly improved performance through large
scaling efforts \citep{zhang2019dialogpt,adiwardana2020meena,boyd2020large,roller2020recipes}, paralleling trends observed in non-conversational text generation \citep{radford2019language,keskar2019ctrl,shoeybi2019megatron}. A challenge of language generation in general, and dialogue in particular,
is that there is more than one valid way
to respond to a given context, e.g., depending
on the unobserved goal of the speaker, the usual tone of their language, their current mood.
Training models over vast amounts of data pooled from millions of users 
with a wide range of opinions and styles means that the resulting generations seem more of a chameleon than of a single, consistent conversational agent.
To address this, researchers have
explored ways to give generation stable grounding (e.g., a persona \cite{zhang2018personalizing,dinan2019second}, knowledge \cite{dinan2018wizard,ghazvininejad2018knowledge}, personal situations \cite{rashkin2019empathy}, internet websites \cite{keskar2019ctrl}, previous conversations from same actor \cite{boyd2020large}), which provide the model with a specific set of talking points, from a potentially huge set. Forms of generation control that are less specific in terms of content and have a much smaller dimension, e.g., sentiment, tone, style, have also been proposed
\citep{keskar2019ctrl,shuster2020image,Dathathri2020Plug}.

In this work, we aim to achieve  control over a medium-sized (217) set of
styles from \citet{shuster2020image},
which still allows for much response variety.
We train a classifier from the data in \citet{shuster2020image}
and show how to adapt three previously proposed promising approaches to this task with large-scale
state-of-the-art conversational models: (1) a retrieve-and-style-transfer approach modified from \citet{weston2018retrieve}, 
  (2) an inference-time iterative refinement procedure proposed in \citet{Dathathri2020Plug}, and
 (3) a conditioned generation approach that fine-tunes the model with augmented
 inputs tagged with the target style, similar to \citet{keskar2019ctrl}.
Comparing trade-offs in terms of performance, cost at training and inference time, and complexity of use, we find that fine-tuned 
conditioned generation yields the
strongest performance, in terms of successful detection of the target style in the generation. Its inference speed is also considerably more tractable compared to the inference-time iterative refinement
procedure.
Automated and human evaluations show that the resulting conversational
models can convincingly match target tones, while largely preserving other conversational metrics. 
This work thus makes the following contributions: (1) we adapt three different approaches for style control to
state-of-the-art conversational architectures
with a mid-size (217) style space
and compare their trade-offs; 
(2) we propose a practical pipeline
combining style-labelled data and unlabelled in-domain conversational data 
that can be generalized to any style space
for which a reasonable classifier can be
trained, and empirically validate that
the resulting model can convincingly
alter the style of conversations without substantially damaging other conversational metrics. Our best style control model and code have been made available through the ParlAI framework;\footnote{\url{https://github.com/facebookresearch/ParlAI/blob/master/parlai/zoo/style_gen/c75_labeled_dialogue_generator.py}} additional models and classifiers mentioned in this paper will be made available soon, also through ParlAI.

The remainder of the paper is organized as follows. Sec.~\ref{sec:related} lists related work. Sec.~\ref{sec:approaches} details the datasets we use and the building blocks of the systems we compare. Sec.~\ref{sec:expe} shows the results of our comparisons. Sec.~\ref{sec:discussion} summarizes the takeaways from this work and proposes future directions.

\section{Related work}
\label{sec:related}

We first present open-domain conversational architectures
that form the foundation of our models, then review previous work for controlling styles.

\subsection{Conversation architectures}

Conversational models can be based on 
generation, retrieval, or a combination of both  (e.g., see \citet{roller2020recipes}). Models including a retrieval component had previously been found to
perform better in human evaluation (e.g., \citet{weston2018retrieve,rashkin2019empathy}).
A recent development has been the dramatic scaling
of transformer-based architectures for language and dialogue generation to billions of parameters 
\citep{radford2019language,keskar2019ctrl,shoeybi2019megatron,adiwardana2020meena,roller2020recipes,boyd2020large,brown2020language}.
Combining such very large models with optimized
beam search, \citet{roller2020recipes}
have obtained higher engagingness and humanness ratings with generative models, compared
to retrieval or retrieve-and-refine approaches \cite{weston2018retrieve}. Approaches that might help smaller models may become moot when higher-capacity models trained on large amounts of data are used instead. In this work,
we use variants of 2.7B-parameter generative models released by \citet{roller2020recipes}, to ensure that the methods work well with state-of-the-art conversational models and produce generations that are as fluent as possible.

\subsection{Styles in conversation}
The evaluation of an open-domain conversational model often relies on asking humans to rate
whether the way the model responds to a given
conversational context is fluent, relevant,
specific, human-like \cite{adiwardana2020meena,roller2020recipes}, which is a fairly minimal set of very generic desirable qualities in generated dialogue. There are many other attributes that could describe a given response, e.g., whether it is polite \cite{niu2018polite}, formal \cite{rao2018dear}, empathetic \cite{rashkin2019empathy}, knowledgeable \cite{dinan2018wizard}, or associated with
a variety of styles \cite{shuster2020image}.
Controlling finer aspects of
generation could enable a more consistent
experience when interacting with a model,
and provide ways to focus on styles that tend to produce generations
with desirable qualities
 such as being pleasant to talk to, less toxic,
 more empathetic, or generally better-behaved \cite{see2019makes,roller2020open,rashkin2019empathy}.
In this work, we use the set of styles proposed in \citet{shuster2019engaging,shuster2020image}, since it has been shown to result in engaging chats when focusing on positive or neutral styles and it is a set of relatively large size (217). \citet{shuster2019engaging} proposes a dataset of images captioned with a target style, while \citet{shuster2020image} provides short conversations with target styles.

\subsection{Controlling style}
\label{sec:controlling_style}
A style control method that has appealing advantages is the plug-and-play method (PPLM) from \citet{Dathathri2020Plug}, an iterative generation method
using a classifier on top of a pre-trained generation model.
For each token, the mean hidden representation of all tokens so far 
is fed into a style classifier. 
A backward pass through the classifier and generator is performed, and the gradients are used to update the activations in the generator's attention layers. These forward and backward passes are repeated 
several times per time step, and the following token is then sampled \citep{Dathathri2020Plug}. 
This work has impressive results and the desirable property that it can be used without having to fine-tune the underlying base model.
We adapt this approach for our purpose, and compare it with conditioned generation approaches.
Another work that is achieving fine-grained control using a very large architecture is the CTRL model \citep{keskar2019ctrl}. The style conditioning relies on control codes obtained from the training data (meta-data). This work is however not tailored to dialogue. 
We previously proposed \citep{lample2018multipleattribute} a style transfer architecture using noisy encoders and decoders and style conditioning through an additional token, and adapted it for use with reasonably large style spaces \citep{smith2019zero}. The style control through an additional context token is similar to the best-performing model in this paper, however the models underlying both these works are much smaller, non-conversational architectures for which generations are considerably less fluent than those of the models we consider here, and the task of rewriting with a given style is more constrained than the conversational generation task that this work focuses on.

\section{Controlling styles with
state-of-the-art conversation architectures}
\label{sec:approaches}

Pieces from previous work 
can be combined into new architectures that can reply to a dialogue context to match a target style. 
We present the conversational datasets we use, then introduce the three
methods we adapt and compare in this work, and their advantages and shortcomings. They differ on whether they use retrieval, and whether
they require fine-tuning of the whole architecture.

\begin{table*}[t!]
    \centering
    \begin{small}
    \begin{tabular}{lp{14cm}}
    \toprule
    \multicolumn{2}{c}{Datasets}\\
    \midrule
IC & Image-Chat:  3-turn conversations about an image, labeled with 217 styles \\
ConvAI2 & ConvAI2: conversations based on personas, no style labels \\
WoW & Wizard of Wikipedia: knowledge-grounded conversations, no style labels \\
ED & EmpatheticDialogues: sentiment-grounded conversations, labeled with 32 sentiments not used in this work \\
BST & BlendedSkillTalk: conversations blending skills from ConvAI2, WoW and ED, no style labels \\
D & non-image-grounded dialogue datasets: ConvAI2 +  WoW + ED + BST, no style labels \\
D+ & D augmented with inferred style labels: ConvAI2 +  WoW + ED + BST, style labels provided by classifier \\
\midrule
\multicolumn{2}{c}{Models}\\
\midrule
         RnST & Retrieve-and-style-transfer \\ 
         PPLM & Plug-and-Play-Language-Model \\
         R & 2.7B generative model pretrained on pushshift.io Reddit \\
         C & Family of conditioned generation models \\
        C0 & Conditioned generation model, no conditioning style token provided during training \\
        C75 & Conditioned generation model, conditioning style token provided for 75\% of training examples \\
        C100 & Conditioned generation model, conditioning style token always provided during training \\
\midrule
\multicolumn{2}{c}{Combinations} \\
\midrule
 C100-IC+D & R fine-tuned on: IC with the style label + D with no style \\
RnST-IC+D & R fine-tuned with retrieved response on: IC with the style label + D with no style \\

    \bottomrule
    \end{tabular}
    \end{small}
    \caption{Shorthand for data and models}
    \label{tab:my_label}
\end{table*}

\subsection{Datasets}
\label{sec:dialogue_datasets}

We use different datasets for providing the style space and fine-tuning most models.

\paragraph{Style space: Image-Chat (IC).} Image-Chat \citep{shuster2020image} is a dataset of 3-turn conversations 
discussing  
an image, totalling about 400k utterances. Each partner in each conversation conveys a given style (e.g., Curious, Casual) from among a set of 217. 
These styles are split into 
``positive,'' ``neutral,'' and ``negative'' (see Table~\ref{tab:imagechat_labels} in the Appendix).
The distribution of styles in the dataset is reasonably balanced and 
the set of styles results in colorful, diverse
conversation \citep{shuster2020image}.
However, this dataset is not a purely conversational dataset because the conversations are referring to an image that both conversation partners are seeing. The dataset can be used to teach a model to condition on a style, but produces conversations that are not self-contained (e.g., "the dog next to the statue seems bored"). Therefore, we also use purely textual datasets to ensure natural conversations without reference to images (see next paragraph). Unfortunately, these textual datasets were collected without providing target styles from the IC style space. We thus
also use Image-Chat to train a classifier which assigns style labels to utterances. We then use this classifier to augment the purely conversational datasets with style labels  (see Sec.~\ref{sec:attribute_labels}).

\paragraph{Dialogue datasets (D).}
Following \citet{roller2020recipes}, we start from models pre-trained on a previously existing Reddit dataset extracted and obtained by a third party and made available on pushshift.io \citep{baumgartner2020pushshift}
, then fine-tune our models on four public open-domain conversational datasets from previous work, collectively denoted by D: (1) \textbf{\conv}, 
comprising about 140k conversations in which two partners discuss themselves 
    based on
    a given  persona, e.g., ``I have four sisters.'' \citep{zhang2018personalizing,dinan2019second};
    (2) \textbf{Wizard of Wikipedia (WoW)}, a dataset of 22k conversations 
    displaying knowledge
    grounded on Wikipedia articles \citep{dinan2018wizard};
    (3) \textbf{EmpatheticDialogues (ED)}, comprising 25k conversations in which one speaker talks about a situation 
    associated with a target sentiment (e.g., Surprised, Proud, Angry),
    and 
    the other 
    responds empathetically \citep{rashkin2019empathy};
    and (4) \textbf{BlendedSkillTalk (BST)}, comprising 5k conversations 
    blending  
    skills of the three previous datasets: conveying a consistent persona, displaying knowledge, and responding empathetically \citep{smith2020can}.

\subsection{Retrieve-and-style-transfer (RnST)}

Based on results in \citet{weston2018retrieve}
showing a superiority of retrieve-and-refine
models over either retrieval or generation
methods for dialogue, 
we considered an 
intuitive approach combining retrieval and
style-controlled generation. 

The best retrieve-and-refine model from \citet{roller2020recipes} first uses a retrieval model to retrieve a response from the training set, then appends that retrieved response to the dialogue context (after a separator),  to generate a
better response. 
Our retrieve-and-style-transfer method (RnST) additionally appends a target style after a second separator.
The model is then trained to generate the gold truth response from that augmented input.
The retriever is far from perfect, which creates
enough noise to prevent the model from
simply copying the retrieved response.
The two elements of (1) noise, and (2) pairing of
a noisy un-styled first guess with a target style
to generate the desired response, are both present
in our recent style transfer
approaches based on added noise and back-translation \citep{lample2018multipleattribute,smith2019zero}.
Another approach that we did not try would
consist in first training a retriever conditioned on style, and then training a vanilla
retrieve-and-refine model using that style-conditioned retriever to provide the first guess.

\subsection{Iteratively refining generations
to match a target style during inference (PPLM)}

The second method (thereafter PPLM) adapts the plug-and-play,
minimal-training approach from \citet{Dathathri2020Plug}
to dialogue and a different set of styles. \citet{Dathathri2020Plug} is based on GPT2, which is not a dialogue model and would therefore be at a disadvantage when evaluated on dialogue. In order to provide a fairer comparison between methods, we replace GPT2
by our C0 model which has been pre-trained on pushshift.io Reddit and fine-tuned on several dialogue datasets (see Sec.\ref{sec:cond_gen}), so that all models we compare share a common base.
We also change the guiding classifier head to accommodate
the style space from Image-Chat.
Given a base model, the PPLM generative method requires no fine-tuning beyond that of the classifier.
Additionally, controlling the style through iterative steps affords direct fine-grained, gradual
control at inference time over the style. Lastly,
the use of a classifier to directly guide the refinement allows to go not only ``towards" a desired style, but also ``away" from it, which
is not straightforward with the other conditioning methods.
However, the inference is also much more costly and might be prohibitive with very large models.
It is also unclear whether the good results 
demonstrated over a small number of classes in text generation would generalize to a much larger set of styles, and to dialogue.

\subsection{Training a conditioned generator on inputs appended with style tags (C)}

The last family of methods that we include in our comparison simply relies on conditioning tokens appended to the dialogue context. We thereafter denote these models by C to reflect their conditioned nature. We fine-tune the 2.7B pushshift.io Reddit pre-trained generative model from \citet{roller2020recipes}, appending target styles to the dialogue context (after a separator). While purely generative models
had long been inferior to retrieval variants in dialogue \citep{weston2018retrieve,rashkin2019empathy}, very recent generative models have been shown
to perform better when combined with beam search
with a minimum output length \citep{roller2020recipes}, making them an attractive base.
This method requires whole-architecture fine-tuning to learn to use the augmented input, but inference is then straightforward. 
Although we do not test this here, fine-grained control over the degree of intensity of the target style 
could be achieved by qualifying the appended style with a degree (e.g., a numerical output from a style classifier capturing intensity of the style of the training example), as in \citet{keskar2019ctrl}, with the limitation
that the degree of control would rely on the available training data and might not directly generalize to other intensities the way the iterative inference PPLM
method promises.

\section{Experiments}
\label{sec:expe}

All experiments are conducted using the ParlAI toolkit \citep{miller2017parlai}. 
In order to fairly compare the approaches in Sec.~\ref{sec:approaches}, we build them as enhancements over the same underlying model. That model is a pushshift.io Reddit pretrained 2.7B model from \citet{roller2020recipes}, thereafter denoted by R, which was pretrained on a previously existing Reddit dataset extracted and obtained by a third party and made available on pushshift.io \citep{baumgartner2020pushshift}. 

\subsection{Use of retrieval}
\label{sec:rnst_results}

As described in the previous section, Retrieve-and-style-transfer combines a retriever and a generator.
Having a retriever take care of finding a relevant response might free up more model capacity to focus on good style conditioning. We do not change the retriever system from \citet{roller2020recipes}, but we modify the generator.

In order to teach the model to condition generations on styles, we fine-tune the generator R on IC with the ground-truth style appended to the dialogue context. However, conversations in IC are all grounded on an image, which is not provided to the model in our architecture (this architecture does not have image features). Fine-tuning solely on IC results in conversations that clearly refer to some unknown image, rather than self-contained conversations. To avoid this problem, we also fine-tune on D, which does not contain style labels.\footnote{No styles from IC were given to workers at data collection time for the datasets in D. ED dialogues were collected with a grounding in a set of 32 sentiments, however these are different from the styles used in IC, and pertain to situations rather than tones as in IC. We do not make use of the sentiment labels from ED in this work. They might be somewhat predictive of IC styles, given that the style spaces are related (e.g., Anxious appears in both sets), but relying on these labels would lead to treating ED differently from the other datasets in D.} 

We compare models fine-tuned either with a retrieved reply appended to the input (RnST-IC+D), or without (C100-IC+D). The C100-IC+D notation captures the fine-tuning on labeled IC and unlabeled D, and the fact that the architecture and training are the same as for C100 in Sec.~\ref{sec:cond_gen}. Full experimental details and an example conversation are given in Appendix ~\ref{sec:appRetNStyle}.

\begin{table}[t!]
\centering
\begin{small}
\begin{tabular}{crr}
\toprule
Model & BST & IC \\
\midrule
RnST-IC+D & 3.3\% & 15.8\% \\
C100-IC+D & 5.7\% & 16.7\% \\
\bottomrule  
\end{tabular}
\end{small}
\caption{\label{tab:rnst}
Accuracies of generations at matching target styles, for a retrieve-and-style-transfer model (RnST-IC+D) and style-conditioned generative model (C100-IC+D), both exposed to IC styles during fine-tuning. Contexts come from either BST or IC. Direct generation has higher accuracies, suggesting that adding a retrieved utterance to the context string does not help generate 
the correct style. Scores are much higher on IC, which is the only dataset with style labels that the models are fine-tuned on.
}
\end{table}

Automatic evaluation of the accuracy of style control is conducted for generation using either IC or BST contexts,
by running a trained style classifier on the model generations and reporting the percentage that get classified into the target style.
The classifier is trained on IC conversations (classifier details given in the Appendix, Sec.~\ref{sec:appClassif}).
Average accuracy on the IC test set itself on turns 2 and 3 is 13.0\% across the 217 classes. This classifier uses both the utterance to be classified and the previous utterance as context (as something might only be, e.g., ``sarcastic'', in the context of what was said before. A classifier using only the utterance itself achieves 12.6\%). 
.

Results in Table \ref{tab:rnst} show that conditioning on retrieved utterances hurts style control. This weaker style control could still be
an acceptable trade-off if the generated reply was of sufficiently higher quality (e.g., more relevant to the dialogue context, which we do not test here),
given the superior results long obtained with retrieval over generation in previous work (e.g., \citet{weston2018retrieve,rashkin2019empathy}). 
However, recent results in \citet{roller2020recipes} have instead
obtained better performance from purely generative models when using larger
architectures and optimized decoding, which we adopted here. Therefore, we expect
that other conversational metrics would also not favor conditioning on a retrieved reply.
A retrieve-and-style-transfer system could still be attractive as a way to 
use one or several out-of-the-box style transfer models without having to fine-tune the whole model for every style space, by simply forming a pipeline from the retriever followed by the style transfer model.

Another observation is that style control is not transferring very well from IC to BST. We also noted when interacting with the models that the image-grounded nature of the IC training conversations resulted in some conversations referring to some unavailable image, which is jarring, even though the model was also fine-tuned on the imageless datasets from D.
In the remainder of this paper, we thus experimented with using IC only
to train the style classifier,and then using that trained style classifier to label D with styles, as detailed in Sec.~\ref{sec:attribute_labels}. We denote by D+ the dataset thus augmented. Once D has been labelled, we fine-tune R exclusively on D+ and drop IC from the fine-tuning step.

\subsection{Labeling D with styles (D+)}
\label{sec:attribute_labels}

The method we outline here provides a way to use an unlabelled dataset with desirable qualities, by leveraging another labelled dataset purely to train a label classifier. Here, the advantage of the other dataset is that it is conversational and self-contained without reliance on an image, but other advantages could be sheer larger magnitude, as in semi-supervised learning.

In practice, we augment each utterance from the four datasets from D with
 style labels, obtained by running the style classifier trained on IC, yielding weakly labeled dataset D+. D+ is used to provide style conditioning during fine-tuning in the remainder of this paper.
\footnote{
A classifier trained on this newly labeled dataset, using only the 
\textit{previous} 
utterance 
as input, obtains 2.1\% accuracy, above the chance level of 0.5\%.  This confirms the intuition
that the previous utterance has some predictive power over the tone of the next utterance in a natural dialogue. This cannot be done on Image-Chat, where the labels were random targets provided to workers instead of organic conversational choices.
}
\begin{table*}[t!]
\small
\centering
\begin{tabular}{lrlrlrlrlr}
\toprule
ConvAI2 & \% & ED (Speaker) & \% & ED (Listener) & \% & WoW & \% & BST & \% \\
\midrule
Curious & 4.2 & Emotional & 4.4 & Sympathetic & 6.9 & Knowledgeable & 17.5 & Curious & 5.2 \\
Businesslike & 3.2 & Appreciative & 3.2 & Questioning & 5.5 & Scholarly & 10.1 & Rustic & 2.9 \\
Youthful & 3.1 & Miserable & 3.0 & Curious & 5.4 & Complex & 6.0 & Questioning & 2.6 \\
Rustic & 2.9 & Anxious & 2.9 & Empathetic & 4.4 & Intelligent & 4.6 & Businesslike & 2.4 \\
Boyish & 2.9 & Resentful & 2.7 & Absentminded & 2.9 & Cultured & 4.2 & Fickle & 2.3 \\
Airy & 2.8 & Sentimental & 2.5 & Optimistic & 2.1 & Obsessive & 1.8 & Open & 2.1 \\
Questioning & 2.5 & Shy & 2.4 & Kind & 2.0 & Rustic & 1.7 & Obsessive & 1.9 \\
Absentminded & 2.5 & Humble & 2.2 & Appreciative & 1.9 & Curious & 1.7 & Knowledgeable & 1.8 \\
Open & 2.1 & Wishful & 2.0 & Excitable & 1.9 & Businesslike & 1.7 & Casual & 1.7 \\
Casual & 1.8 & Fickle & 1.8 & Bewildered & 1.8 & Meticulous & 1.7 & Sympathetic & 1.6 \\
Maternal & 1.7 & Optimistic & 1.7 & Caring & 1.6 & Passionate & 1.4 & Passionate & 1.6 \\
Relaxed & 1.6 & Businesslike & 1.6 & Compassionate & 1.5 & Brilliant & 1.4 & Youthful & 1.6 \\
\bottomrule  
\end{tabular}
\caption{\label{tab:top_attributes} 
The most common styles, by percentage frequency, found in the training sets of several dialogue tasks (ConvAI2, EmpatheticDialogues, Wizard of Wikipedia, and BlendedSkillTalk), according
to our style classifier.  
For each dataset, the top styles reflect the type of dialogue that that dataset was designed to demonstrate: curiosity and openness for ConvAI2, emotion and expressiveness from the Speakers in EmpatheticDialogues, sympathy and inquisitiveness from the Listeners in EmpatheticDialogues, knowledge and intelligence 
in Wizard of Wikipedia, and a blend of the above in BlendedSkillTalk.
}
\end{table*}
The empirical distribution of style types fed to models during training consists of 51\%
positive, 20\% neutral, and 29\% negative styles (see 
Table~\ref{tab:imagechat_labels} in the Appendix for a breakdown of style types).

The top 12 styles 
in each of the datasets in D+ are shown in Table~\ref{tab:top_attributes}  (utterances of ED are split by whether they were said by the Speaker, who talks about an emotional situation, or the Listener, who  
responds with empathy). 
Top styles show patterns that reflect the intended focus of each dataset.
For instance, the ConvAI2 dataset 
instructed workers to get to know each other 
 (``Curious'', ``Questioning'', ``Open''); the Speakers in ED 
 were instructed to talk about emotional situations 
 (``Emotional'', ``Appreciative'',``Miserable'', ``Anxious'', etc), and the Listeners to respond with empathy (``Sympathetic'', ``Empathetic'', ``Kind'', ``Caring'', etc); and the WoW utterances were grounded on knowledge from Wikipedia
 (``Knowledgeable'', ``Scholarly'', ``Intelligent'', etc). The BST dataset, designed to incorporate the conversational skills of the above three datasets, contains top styles from all of them (``Open'' from ConvAI2, ``Curious'' and ``Questioning'' from ConvAI2 and ED, ``Fickle'' and ``Sympathetic'' from ED, ``Knowledgeable'' and ``Obsessive'' from WoW, etc.). 
 This provides some empirical validation of the intended focus of each of these datasets, and shows
 that the trained classifier can usefully tease apart styles that correlate with specific conversational qualities, despite the overall relatively low accuracy on IC itself.

\subsection{Conditioned generator fine-tuning (C)}
\label{sec:cond_gen}

We fine-tune R on D+, with a kind of "style drop-out:" 
the style label of each example 
is sometimes joined to the end of the example's context with a \texttt{STYLE} string,
similar to the conditioning in \citet{weston2018retrieve}. 
Starting from the same pre-trained model,
we fine-tune three versions, C0, C75, and C100, 
which are given the appended style for 0\%, 75\%, and 100\% of the training examples, respectively.
We generate with beam search with the best setting from \citet{roller2020recipes}. We do not alter the natural empirical
distribution of styles in D+ (e.g., by upsampling under-represented styles)
in order to better match natural unconstrained dialogue; however, 
upsampling could be used for better performance on less frequent styles.
Appendix~\ref{sec:appContext} and \ref{sec:appCond} give more details. A random sample of generations is shown in Table~\ref{tab:random_grouped_gens}, with many more generations shown in Appendix ~\ref{sec:appModGen}. The examples shows that
style can be controlled with a clear differentiation between different styles, while keeping the responses both fluent and relevant to the dialogue context. As for what the "style" qualitatively captures, it appears to be a mixture of persona traits, moods, tones, and word choice biases.

\begin{table}[t!]
\centering
\begin{small}
\begin{tabular}{p{7.2cm}}
\toprule
\textit{You can't love someone else until you love yourself} \\
\textbf{[Wishful]} I wish I could love myself, but I don't know if I will ever be able to. \\
\textbf{[Fanatical]} That's so true! My mom is a nurse and she always tells me the same thing! \\
\textbf{[Cowardly]} I don't think I could ever be in a long term relationship because I'm too afraid of commitment. \\
\textbf{[Frightening]} That's so true! You have to be happy with yourself before you can be happy in a relationship. \\
\midrule
\textit{It's hard work, but I get a stipend for teaching labs for the department. :) So that's cool. I'm an English major both undergrad and MA.} \\
\textbf{[Rustic (Rural)]} English is a great major. I grew up in the south so I know a lot of English majors. \\
\textbf{[Angry]} That's great! I hate to say it, but English is one of my worst subjects! \\
\textbf{[Scholarly]} Psychology is a very interesting subject. It is the science of mind and behavior.  \\
\textbf{[Earnest (Enthusiastic)]} That's great! English is one of my favorite subjects. How long have you been studying English? \\
\midrule
\textit{It was great. I enjoyed eating it while listening to Bach,} \\
\textbf{[Businesslike]} I love Bach. What do you do for a living? I work at a bakery. \\
\textbf{[Knowledgeable]} Bach is a great composer. He was born in Bavaria, Germany. \\
\textbf{[Open]} That sounds like a great way to spend the evening. What kind of music do you like? \\
\textbf{[Happy]} That's great! I love listening to music while I eat. What kind of cake was it? \\
\midrule
\textit{Yes! I'd better get to the store soon to pick up some salt. 
When do you guys normally start seeing snow around here?} \\
\textbf{[Caring]} I'm not sure, but I hope it's not too late in the year for it to start. \\
\textbf{[Sympathetic]} I think it depends on where you are. I live in the south so we don't get much snow. \\
\textbf{[Scholarly]} It depends on where you are. In the Midwest, it can start as early as January. \\
\textbf{[Appreciative (Grateful)]} I'm not sure, but I'm glad I don't have to deal with it. \\
\midrule
\textit{you flatter me! what hobbies do you participate in?} \\
\textbf{[Sympathetic]} I don't have many hobbies, but I do volunteer at an animal shelter. \\
\textbf{[Open]} I like to volunteer at the animal shelter. What about you? Do you have any hobbies? \\
\textbf{[Scholarly]} I like to read and write poetry. What about you? What do you like to do? \\
\textbf{[Vacuous (Empty, Unintelligent)]} I don't really have any hobbies. What do you do for a living, if anything? \\
\bottomrule  
\end{tabular}
\end{small}
\caption{\label{tab:random_grouped_gens}
Randomly selected set of generations from model C75. 
For each context (italics, from the BST test set), we show responses generated conditioned on four random target styles (bold).
The generations show clearly differentiated styles in the response that are still adequate in context.
For simplicity, only the last utterance of the context is shown, but the model had access to longer contexts. Many more randomly selected generations are shown in Appendix ~\ref{sec:appModGen}. 
}
\end{table}

\subsection{PPLM inference}
\label{sec:pplm_eval}

\citet{Dathathri2020Plug} exclusively presents results on a binary sentiment generation task for demonstrating how PPLM can steer GPT2, using very positive and very negative classes trained on movie reviews from the SST5 dataset \citep{socher-etal-2013-recursive}.
In order to check that our implementation performs similarly to the original implementation in \citep{Dathathri2020Plug}, we first run experiments using that 2-class sentiment
space. 
We then run experiments with our space of 217 IC styles.

The PPLM approach requires a generative model to plug in, with a classifier head on top. 
We use R fine-tuned on unlabelled data in the relevant domain space -- i.e., on SST5 when working on the binary sentiment generation task, or on D when working in the space of open-domain conversations.
The classifier head is a linear layer with an input dimension of 2560, and as many output units as classes, fine-tuned either on SST5 or on turns 2 and 3 of IC, with the 
decoder output
averaged across time 
as in \citet{Dathathri2020Plug}.
We also fine-tune C75 on SST5 for comparison in the SST5 space (C75-SST5).
Additional details and more extensive results are given in Appendix~\ref{sec:pplm_details}.

Table \ref{tab:pplm_resources} shows that the PPLM approach is much more attractive in terms of resource requirements at fine-tuning time, especially for the binary SST5 space, with much faster convergence and lower memory use.
Table~\ref{tab:pplm_concise} shows generation times and percentages of the time when the generation of the model is classified as having the target style.  \citet{Dathathri2020Plug} measure accuracies of matching the target style for SST5 using an external classifier fine-tuned on the Large Movie Review Dataset \citep{maas-EtAl:2011:ACL-HLT2011}. Therefore this section provides our experimental results using a classifier fine-tuned on that same dataset, solely for comparison purposes.
When conditioning generation on SST5 movie-review ratings, our PPLM results are comparable to the
accuracy in \citet{Dathathri2020Plug}, while our C75 results are slightly above.
In the larger space of styles from the Image-Chat dataset, PPLM inference results in accuracies closer to chance and considerably longer inference time.

Based on this performance differential for our style space and base models, we only consider our C models in the rest of the paper.

\begin{table}[h!]
\centering
\begin{small}
\begin{tabular}{lrrr}
\toprule
Tuning & GPUs & GPU mem & Convergence \\
\midrule
C75-SST5 & 8 & 55\% & 10.5 hr \\
PPLM-SST5 & 8 & 7\% & 0.4 hr \\
\midrule
C75-IC & 8 & 73\% & 9.6 hr \\
PPLM-IC & 8 & 11\% & 13.5 hr \\
\bottomrule  
\end{tabular}
\end{small}
\caption{\label{tab:pplm_resources}
Resource use for fine-tuning the C75 model or the top PPLM classifier head on SST5 or IC.
The PPLM classifier head requires much fewer GPU-memory-hours, especially with SST5 where the number of classes is very small.
}
\end{table}

\begin{table}[h!]
\centering
\begin{small}
\begin{tabular}{ccrrrr}
\toprule
Styles & Model & Acc & GPUs & Mem & Gen time \\
\midrule
SST5 & C75 & 82.2 & 1 & 65\% & 3.0 s \\
& PPLM & 76.7 & 1 & 65\% & 69.4 s \\
\midrule
IC & C75 & 7.1 & 1 & 65\% & 1.7 s \\
& PPLM & 1.7 & 1 & 65\% & 45.6 s \\
\bottomrule  
\end{tabular}
\end{small}
\caption{\label{tab:pplm_concise}
Automatic metrics for generating from C75 or our PPLM.
On the SST5 task, 
accuracy for our PPLM is similar to that
reported in \citet{Dathathri2020Plug}, while C75 is slightly above.
However, in the much larger IC style space,
PPLM is much closer to the 0.5\% chance level than C75. GPU memory usage is the same in all cases.
Generation time is
a lot faster for C75, reflecting a different trade-off between investment at fine-tuning time or test time.
}
\end{table}

\subsection{Automated metrics evaluation for C}
\label{sec:auto_eval}

\begin{table}[t!]
\centering
\begin{small}
\begin{tabular}{cccrrrr}
\toprule
& & & \multicolumn{2}{c}{Train dist} & \multicolumn{2}{c}{Uniform dist} \\
\cmidrule(lr){4-5} \cmidrule(lr){6-7}
Task & Model & Cond & Acc & PPL & Acc & PPL \\
\midrule
BST & C100 & + & 27.8 & 4.26 & 14.0 & 4.33 \\
& C100 & - & 1.3 & 4.03 & 0.4 & 4.03 \\
& C75 & + & 23.2 & 4.17 & 11.7 & 4.23 \\
& C75 & - & 1.3 & 3.76 & 0.5 & 3.76 \\
& C0 & - & 1.4 & 3.63 & 0.5 & 3.63 \\
\midrule
IC & C100 & + & 31.6 & 4.92 & 18.4 & 5.03 \\
& C100 & - & 0.9 & 4.86 & 0.5 & 4.86 \\
& C75 & + & 29.3 & 4.97 & 17.2 & 5.07 \\
& C75 & - & 0.9 & 4.65 & 0.5 & 4.65 \\
& C0 & - & 1.1 & 4.50 & 0.5 & 4.50 \\
\bottomrule  
\end{tabular}
\end{small}
\caption{\label{tab:gens}
Style accuracies (i.e., agreement between classifier and target style) and perplexities of model generations.  
Generations were produced using the conversation histories of the BlendedSkillTalk (BST) and Image-Chat (IC) test sets as contexts.
We show results for three models, which were
provided with conditioning style labels 100\%, 75\%, and 0\% of the time during fine-tuning. 
``Cond'' specifies whether each model is conditioned on style labels for the generations being scored. 
Target labels for generations were distributed either according to the distribution that the models were fine-tuned on (``Train dist'') or uniformly. The model conditioned on style labels during fine-tuning 75\% of the time (C75) has generations whose style accuracies approach that of the model conditioned on style labels 100\% of the time (C100), but in addition, when no style conditioning is used, the C75 model creates generations with perplexities that are lower than those of the C100 model and closer to the baseline model that was never conditioned on style labels during fine-tuning (C0).
}
\end{table}

Table~\ref{tab:gens} displays the accuracies of C models' generations at matching target styles, 
and the perplexities of those generations. We test generations with contexts from the test sets of both BST and IC, and for each generation we condition on one of the IC styles present in the training set of D+.
\footnote{See Table~\ref{tab:imagechat_labels} in the Appendix for more details.} We choose the distribution of target styles for these generations in two ways: matching the empirical distribution of styles that the models were fine-tuned on, and uniformly across all styles. For both distributions, we produced roughly 21,500 generations, or roughly 100 generations per target style on average. 

\paragraph{Style control.} For C75 and C100 conditioned on style, accuracies of matching the target style range from 23\% to 32\% on the training distribution of styles and from 11\% to 19\% uniformly across styles. C0 performs at chance on the uniform distribution, and a bit over chance when following the empirical style distribution.
Note that the 11.7\% accuracy result on BST for C75 tested on a
uniform style distribution differs from the 7.1\% result in the comparison with PPLM (Table~\ref{tab:pplm_concise}). The generation settings are different: in particular, in the comparison with PPLM, the first few words of the generated text are directly copied from the gold response, which has a random, arbitrary style (and not the target style), before the model starts generating from C75.

\paragraph{Perplexity of generations.} For the C100 and C75 models, we report accuracies and perplexities both with and without style conditioning during generation. Perplexities of generations were computed using a separate 90M-parameter generative model pretrained on pushshift.io Reddit and fine-tuned on the four dialogue datasets listed in Section~\ref{sec:dialogue_datasets} \citep{roller2020recipes}. 
Perplexity gets slightly worse from training with style conditioning, but this effect is mitigated by the style drop-out used for training C75, for which perplexities are very close to C0 when no style conditioning is used.

\begin{table}[t!]
\centering
\begin{small}
\begin{tabular}{ccrrrr}
\toprule
& & \multicolumn{4}{c}{Utterances used to predict style} \\
\cmidrule(lr){3-6}
Model & Cond & (none) & Prev & Curr & Prev + curr \\
\midrule
C100 & + & 10.29 & 11.30 & 9.39 & 9.34 \\
C75 & + & 10.15 & 11.14 & 9.49 & 9.44 \\
C0 & - & 9.93 & - & - & - \\
\bottomrule  
\end{tabular}
\end{small}
\caption{\label{tab:ppls_with_our_model}
Perplexities of the BST test set as measured by three of the models in Table~\ref{tab:gens}, where ``Cond'' indicates whether the model was allowed to condition on any style labels during inference. We report perplexities for four different cases, depending on whether the conditioning style provided to the model was predicted from the target utterance (``Curr''), the previous utterance (``Prev''), both, or whether we applied no style labels at all. 
Conditioning on styles predicted using the target utterance, or the target utterance and the previous utterance, lowers perplexity for models trained to condition on styles.
However, the increased perplexity when conditioning
on an style predicted from the previous utterance alone (using a classifier trained to infer the style of the following line in our dialogue datasets) suggests that this prediction is too noisy to be used alone.
The very slightly lower perplexity when using both the previous and current utterances to predict labels, compared to only the current utterance, is consistent with the very slightly higher accuracy of the classifier that uses both instead of just the current utterance for prediction on Image-Chat.
}
\end{table}

\paragraph{Perplexity of BST test set.} To gauge whether predicting the style of an utterance can help with generation, we compare perplexities of the BST test set as measured by our models, as a function of whether 
generation is conditioned on a label, and what classifier was used to produce that label. Results are shown in Table~\ref{tab:ppls_with_our_model}. 
For the Previous utterance result, we fine-tune R on D+
\footnote{The styles used to train this classifier were obtained as described in Sec.~\ref{sec:attribute_labels}} 
using only the \textit{previous} utterance as context. BST test-set examples labeled with the styles predicted with this classifier have higher perplexities than using no styles at all, 
reflecting the fact that a single utterance is only a weak predictor of the style of the following utterance. However, perplexities are lower when style labels are predicted using classifiers trained on (utterance, style) pairs from turns 2 and 3 of Image-Chat, implying that these style labels convey meaningful information about the utterances.
The perplexities drop slightly lower 
the classifier uses both current and previous utterance, indicating that the previous utterance may contain a bit of contextual information that is useful for predicting the appropriate style label.

\subsection{Human evaluation}
\label{sec:human_eval}

\begin{table}[t!]
\centering
\begin{small}
\begin{tabular}{ccrrrrr}
\toprule
Model & Cond & Acc & Emp & Rel & Hum & Eng \\
\midrule
C100 & + & 41.3 & 3.93 & 4.03 & 3.86 & 3.87 \\
C75 & + & 34.9 & 4.00 & 4.23 & 3.77 & 4.00 \\
C75 & - & 18.2 & 4.12 & 4.20 & 4.10 & 4.08 \\
C0 & - & 14.2 & 4.09 & 4.12 & 4.06 & 4.04 \\
\bottomrule  
\end{tabular}
\end{small}
\caption{\label{tab:human_eval}
Human evaluations of our models. Evaluators were asked to converse with our models and then try to guess the style that that model was conditioned on out of a set of 5 choices (Acc). They were also asked to rate from 1 to 5 how empathetic, relevant, human-like, and engaging the model's responses were. Evaluators are much more likely to identify the correct styles for the models conditioned on styles during generation, at the cost of those responses being somewhat less human-like. Models were conditioned on 5 common ``positive'' style labels and 5 common ``negative'' style labels.
}
\end{table}

Table~\ref{tab:human_eval} gives the results of crowdsourced human ratings of our models. In line with our automated metrics from Section~\ref{sec:auto_eval} showing our models' ability to effectively use style labels during generation, evaluators correctly identify the target style of our models 34\% to 42\% of the time when the model is conditioned on that style label, but only 14\% to 19\% of the time when the style label is not used during generation. Scores on other metrics (empathy, relevance of response, humanness, and engagingness) are largely unchanged when conditioning on styles or not, except for humnanness, which decreases somewhat when conditioning. 
Accuracy differences are statistically significant for every possible pairing of an style-conditioned and an unconditioned model. The difference in humanness score between C75 with and without conditioning is significant, as is the difference in humanness between C75 with conditioning and C0 without. All other differences are not significant. Additional experimental details can be found in Section~\ref{sec:human_eval_details} of the Appendix.

\section{Discussion}
\label{sec:discussion}

This work explored ways to combine state-of-the-art open-domain conversational architectures with style control for a reasonably large set of styles (217). These methods have different advantages. The retrieve-and-style-transfer approach we tried yielded weaker style control compared to conditioned generation without retrieval, however combining retrieval with style transfer would allow to use out-of-the-box style transfer methods without fine-tuning and transfer into many different style spaces.
The PPLM-style approach is considerably cheaper at train time, however it does not perform very well for larger style spaces, and inference is a lot slower. The conditioned generation approaches we tested can convincingly generate sets of varied 
conversational replies that display the desired style, with barely any cost in terms of other conversational metrics, as shown through automatic and human evaluation, and evident in sample generations.
While we focused on a specific set of styles, our approach should generalize to any set of styles for which a classifier is available, by following
the procedure of labeling dialogue datasets with
the classifier and fine-tuning on that weakly labeled set.
Future work will extend this approach to unsupervised style spaces and styles directly inferred from a conversational partner.
Another promising direction is to investigate
 whether certain utterance-level style
trajectories in conversations are particularly appealing in a conversational agent or to maximize a specific conversational goal,
for example by using reinforcement learning
techniques to learn optimal policies in the space
of style sequences.

\bibliographystyle{acl_natbib}
\bibliography{emnlp2020}

\newpage
\onecolumn
\appendix
\input{appendix}

\end{document}

%% file: appendix.tex
\section{Image-Chat styles}
\label{sec:labels}
The partition of Image-Chat styles by type is given in Table~\ref{tab:imagechat_labels}.
\begin{table}[h!]
\small
\centering
\begin{tabular}{p{15.5cm}}
\toprule
\textbf{Positive}: Adventurous, Appreciative (Grateful), Articulate (Well-spoken, Expressive), Attractive, Brilliant, Calm, Captivating, Caring, Charming, Cheerful, Clever, Colorful (Full of Life, Interesting), Compassionate (Sympathetic, Warm), Confident, Considerate, Contemplative (Reflective, Thoughtful), Courageous, Creative, Cultured (Refined, Educated), Curious, Daring, Deep, Dramatic, Earnest (Enthusiastic), Elegant, Eloquent (Well-spoken, Expressive), Empathetic, Energetic, Enthusiastic, Exciting, Extraordinary, Freethinking, Fun-loving, Gentle, Happy, Honest, Humble, Humorous, Idealistic, Imaginative, Insightful, Intelligent, Kind, Knowledgeable, Logical, Meticulous (Precise, Thorough), Objective (Detached, Impartial), Observant, Open, Optimistic, Passionate, Patriotic, Peaceful, Perceptive, Playful, Practical, Profound, Rational, Realistic, Reflective, Relaxed, Respectful, Romantic, Rustic (Rural), Scholarly, Sensitive, Sentimental, Serious, Simple, Sophisticated, Spirited, Spontaneous, Stoic (Unemotional, Matter-of-fact), Suave (Charming, Smooth), Sweet, Sympathetic, Vivacious (Lively, Animated), Warm, Wise, Witty, Youthful \\
\midrule
\textbf{Neutral}: Absentminded, Aggressive, Amusing, Artful, Boyish, Breezy (Relaxed, Informal), Businesslike, Casual, Cerebral (Intellectual, Logical), Complex, Conservative (Traditional, Conventional), Contradictory, Cute, Dreamy, Dry, Emotional, Enigmatic (Cryptic, Obscure), Formal, Glamorous, High-spirited, Impersonal, Intense, Maternal (Mother-like), Mellow (Soothing, Sweet), Mystical, Neutral, Old-fashioned, Ordinary, Questioning, Sarcastic, Sensual, Skeptical, Solemn, Stylish, Tough, Whimsical (Playful, Fanciful) \\
\midrule
\textbf{Negative}: Abrasive (Annoying, Irritating), Airy (Casual, Not Serious), Aloof (Detached, Distant), Angry, Anxious, Apathetic (Uncaring, Disinterested), Argumentative, Arrogant, Artificial, Assertive, Barbaric, Bewildered (Astonished, Confused), Bizarre, Bland, Blunt, Boisterous (Rowdy, Loud), Childish, Coarse (Not Fine, Crass), Cold, Conceited (Arrogant, Egotistical), Confused, Contemptible (Despicable, Vile), Cowardly, Crazy, Critical, Cruel, Cynical (Doubtful, Skeptical), Destructive, Devious, Discouraging, Disturbing, Dull, Egocentric (Self-centered), Envious, Erratic, Escapist (Dreamer, Seeks Distraction), Excitable, Extravagant, Extreme, Fanatical, Fanciful, Fatalistic (Bleak, Gloomy), Fawning (Flattering, Deferential), Fearful, Fickle (Changeable, Temperamental), Fiery, Foolish, Frightening, Frivolous (Trivial, Silly), Gloomy, Grand, Grim, Hateful, Haughty (Arrogant, Snobbish), Hostile, Irrational, Irritable, Lazy, Malicious, Melancholic, Miserable, Money-minded, Monstrous, Moody, Morbid, Narcissistic (Self-centered, Egotistical), Neurotic (Manic, Obsessive), Nihilistic, Obnoxious, Obsessive, Odd, Offhand, Opinionated, Outrageous, Overimaginative, Paranoid, Passive, Pompous (Self-important, Arrogant), Pretentious (Snobbish, Showy), Provocative, Quirky, Resentful, Ridiculous, Rigid, Rowdy, Scornful, Shy, Silly, Stiff, Stupid, Tense, Uncreative, Unimaginative, Unrealistic, Vacuous (Empty, Unintelligent), Vague, Wishful, Zany \\
\bottomrule  
\end{tabular}
\caption{\label{tab:imagechat_labels}
Styles in the Image-Chat training set, separated by ``positive'', ``neutral'', and ``negative'' categories, as listed in \url{http://ideonomy.mit.edu/essays/traits.html}. 215 styles are listed; 2 additional attributes, ``Crude'' and ``Earnest'', appear only in a handful of Image-Chat training set examples and are included among our classifier's 217 classes, but do not appear in any of the attribute labels that the classifier attached to our four dialogue datasets. Generative models are fine-tuned on the 215 different attribute labels that were applied to at least one utterance of our four dialogue datasets.
}
\end{table}

\section{Retrieve-and-style-transfer architecture and training}
\label{sec:appRetNStyle}

\begin{table}[h!]
\small
\centering
\begin{tabular}{p{15.5cm}}
\toprule
Human: Hi there! \\[1.5ex]
\textit{(Retriever: Hey hey!)} \\
Generator: \textbf{[Honest]} I'm not sure what you mean by that, but I'm sure you're a nice person. \\[1.5ex]
Human: That's nice of you to say! Have you had a good weekend? \\[1.5ex]
\textit{(Retriever: My weekend was pretty okay. Did anything special happen for you?)} \\
Generator: \textbf{[Rustic (Rural)]} My weekend was pretty okay. Did anything special happen for you? (I live in the country) \\[1.5ex]
Human: Not really! Walked around a bit. That's cool that you live in the country - where at? \\[1.5ex]
\textit{(Retriever: Upstate, where I grew up. You?)} \\
Generator: \textbf{[Serious]} I live in Georgia. It's a pretty nice place to live. What about you? \\
\bottomrule  
\end{tabular}
\caption{\label{tab:rst_convo}
Start of a randomly chosen sample conversation with the retrieve-and-style-transfer generator model evaluated in Section \ref{sec:rnst_results}. Each of the generator's responses is conditioned on a different randomly selected attribute label (bold). Retrieved utterances, appended to contexts alongside target attribute labels, are shown in italics. Generator sometimes effectively generates using the target attribute and sometimes copies the retrieved utterance.
}
\end{table}

The retrieve-and-style-transfer architecture we use is the retrieve-and-refine architecture from \citet{roller2020recipes}, but fine-tuning with Image-Chat examples with their style tag.
The architecture 
consists of (1) a retriever model used to select an appropriate response given candidates, and (2) a generator model in which the retriever's response and the attribute of the gold response are appended to the context string during training. The retriever model is a 660M-parameter Poly-encoder, consisting of two Transformer encoders for context strings and candidate responses, whose outputs are attended over to produce a ranking of candidates \citep{humeau2020poly}. The model has 24-layer encoders, 16 attention heads, an embedding size of 1024, a feed-forward size of 4096, and 64 Poly-encoder context codes. The model is pretrained on a previously existing third-party Reddit dump that was hosted by pushshift.io \citep{baumgartner2020pushshift} and fine-tuned on ConvAI2, ED, WoW, BST, and turns 2 and 3 of Image-Chat. For ConvAI2, ED, and WoW, we fine-tune on versions of the datasets to which persona strings (like in ConvAI2) and conversational topics (like in WoW) have been added if they are not already present, as in \citet{smith2020can}. This is done to better match the contexts of these three datasets to each other and to those of BST, which includes persona strings and often WoW topics in its contexts.

To fine-tune the retriever, we tune both the learning rate and the relative training weights of the datasets, and we use accuracy at retrieving the gold response as our validation metric. After retriever fine-tuning, we cache retriever responses for all datasets that we wish to fine-tune our generator on, to speed up generator fine-tuning.

Our generator model uses the same architecture as in \ref{sec:cond_gen}. During training, the cached retriever response for each example is joined to the end of the context string with a ``\texttt{ RETPRED }'' string, and the attribute of the gold response is joined to the end of that with a ``\texttt{ STYLE }'' string, similar to how retriever responses are handled in \citet{weston2018retrieve}. We fine-tune the generator on the same five datasets as with the retriever; however, among the datasets only Image-Chat contained attribute labels, and so the model did not see attribute strings appended to contexts when training on examples from the other four datasets.

To fine-tune the generator, we sweep the learning rate and the relative training weights of the datasets, as well as the fraction of the time that the gold response is appended to the context in place of the retrieved response in order to teach the generator to sometimes copy that response. During generation, candidates from ConvAI2, ED, WoW, and BST are ranked by the retriever, and the top retrieval candidate and target attribute are appended to the end of the context. An example conversation from our retrieve-and-style-transfer generator is shown in Table~\ref{tab:rst_convo}.

\section{Attribute classifier architecture}
\label{sec:appClassif}
The attribute classifier trained on the Image-Chat attribute space consists of R (a 2.7B-parameter Transformer model pre-trained on a previously existing third-party Reddit dump that was hosted by pushshift.io \citep{baumgartner2020pushshift} from \citet{roller2020recipes}), with an added 
linear layer with a hidden dimension of 2560 
on top of the decoder output. 
We fine-tune all weights 
on turns 2 and 3 of the Image-Chat training set, using the provided labels. (We do not train on turn 1, which relies more centrally on the image, according to \citet{shuster2020image}).

\section{Conversation context given to the model during dialogue fine-tuning}
\label{sec:appContext}
For ConvAI2, ED, and WoW, we fine-tune on versions of the datasets in which persona strings and conversational topics have been added to all contexts, as in Section \ref{sec:appRetNStyle}. These contexts are better matches to the contexts used in the human evaluations of Section~\ref{sec:human_eval}, in which two persona strings are assigned to both the human and to the bot during conversation. During training, examples are sampled from the ConvAI2, ED, WoW, and BST datasets with a ratio of 1:2:1:1, adopted from models trained on these datasets in \citet{smith2020can}.

\section{Training the style-conditioned model}
\label{sec:appCond}
We fine-tune 3 models, for which the style label is randomly appended to the context string 100\%, 75\%, and 0\% of the time during training. For each training example, a random number in the unit interval is drawn to determine whether to append that example's style label to its context string, given the specified probability. The 0\%-probability model (C0) serves as a baseline for the 100\%-probability model (C100), and the 75\%-probability model (C75) allows for generation in which an style label is useful but not required, because the C75 model has been exposed to both cases during fine-tuning. Models were trained with a batch size of 128 and 8 GPUs, and the learning rate was swept in the range of 3e-6 to 7e-5, with perplexity used as validation metric. The C100 model converged in 8.7 hours, the C75 in 9.6 hours, and the C0 in 22.5 hours; however, the C0 model had a slightly lower learning rate, which likely resulted in the longer training time.

\subsection{Generation parameters}
\label{sec:appGen}
For style-controlled generation with our fine-tuned models, we use beam search with a beam size of 10, a minimum beam length of 20, and $n$-gram blocking of size 3 in both the beams and the context, following \citet{roller2020recipes}. Generations take roughly 2.0 seconds per generation, with a batch size of 32 across 4 GPUs, and generation speeds are roughly equivalent with and without style conditioning.

\section{PPLM comparison}
\label{sec:pplm_details}

\subsection{Experiment with SST-5 attributes}
\label{sec:pplm_sst5_experiment_details}

The metrics and evaluation datasets in this section follow \citet{Dathathri2020Plug}.
Since the SST-5 dataset \citep{socher-etal-2013-recursive} consists of review/rating pairs without any context strings, only a ``\texttt{\_\_SILENCE\_\_}'' string is passed into the encoder during fine-tuning of the generators and during classifier-head tuning. The 15 2-to-5-word prefixes in \citet{Dathathri2020Plug} are used at the beginnings of generations, as was done in that work. The learning rate is swept from 2e-6 to 3e-5 during generator fine-tuning and from 2e-3 to 3e-1 during classifier-head tuning. Like \citet{Dathathri2020Plug}, we pick tokens at each timestep for C75 and PPLM by sampling the token distribution with top-$k$ filtering ($k=10$, \citet{fan2018hierarchical}); unlike \citet{Dathathri2020Plug}, however, we stop a generation when it hits an end-of-sentence token, as in \citet{roller2020recipes}. For PPLM, we find that varying the step size of gradient updates leads to a trade-off between increased attribute control and degeneration of the output utterances; we tune the step size in the range of 0 to 0.1 (where step size is defined by $\alpha$ in \citet{Dathathri2020Plug}), and we find that a step size of 0.07 leads to maximum average accuracy of the target attribute. Final numbers come from re-running generations with that step size and a different seed.

\begin{table}[t!]
\centering
\begin{small}
\begin{tabular}{lrrrr}
\toprule
Method & Acc (\%) & Dist & \# tokens & Gen time (s) \\
\midrule
\multicolumn{2}{l}{\textit{C75:}} \\
B & 82.2 & 0.82 & 28.8 & 3.0 \\
B* & 83.3 & 0.89 & 20.9 & 2.1 \\
\midrule
\multicolumn{2}{l}{\textit{PPLM:}} \\
B & 50.0 & 0.86 & 25.6 & 57.9 \\
B* & 50.0 & 0.88 & 23.8 & 53.1 \\
BR & 73.3 & 0.87 & 23.4 & 52.3 \\
BC & 66.7 & 0.83 & 33.8 & 71.6 \\
BCR & 76.7 & 0.89 & 33.4 & 69.4 \\
\bottomrule  
\end{tabular}
\end{small}
\caption{\label{tab:pplm_sst5}
Automatic metrics for one of our style-conditioned models (C75) vs. a model (PPLM) in which a classifier head has been trained on ``very positive'' and ``very negative'' classes of the movie-review dataset SST-5 \citep{socher-etal-2013-recursive} and used for iterative inference as in  \citet{Dathathri2020Plug}. Both models are fine-tuned on 
that SST-5 data
(in the case of PPLM, before tuning the classifier head). We report accuracy at generating the target attribute, the mean of the Dist-1, Dist-2, and Dist-3 scores of distinct $n$-grams (Dist), the mean number of total tokens per generation, and the mean number of seconds per generation. \textbf{B}: baseline generation, sampled once; \textbf{B*}: the first remaining generation of a group of 10, after filtering out all those with a Dist score below 0.75; \textbf{BR}: the generation of a group of 10 with the lowest classifier loss, after Dist filtering; \textbf{BC}: a generation sampled once, using the PPLM technique of updating latent representations; and \textbf{BCR}: the generation of a group of 10, created with the PPLM technique, that has the lowest classifier loss, after Dist filtering. Each generation is prefixed with one of a list of 15 different phrases, from \citet{Dathathri2020Plug}. 
Generations from C75 are classified as having the target attribute more often than PPLM-style models, with considerably faster generation times. The accuracies of the PPLM model are comparable to accuracies reported in \citet{Dathathri2020Plug}.
}
\end{table}

More complete results comparing our C75 and PPLM models, using the SST-5 movie-reviews dataset, are shown in Table~\ref{tab:pplm_sst5}. We report metrics under different experimental conditions, taken from \citet{Dathathri2020Plug} (with the exception of \textbf{B*}):

\begin{itemize}
    \item \textbf{B}: take the mean over 10 generations for each target attribute (implemented here by ``perturbing'' attention activations with a step size of 0)
    \item \textbf{B*}: produce 10 groups of 10 generations for each target attribute. For each generation, calculate the mean of the Dist-1, Dist-2, and Dist-3 scores, which measure token diversity \citep{li2015diversity}; throw out the generation if this mean (Dist) is below a certain threshold (here, 0.75, in order to retain at least one generation per group\footnote{\citet{Dathathri2020Plug} used a threshold of 0.9 to filter generations by Dist score. One hypothesis for why our generations tended to have lower Dist scores is because our generations' average token length is much shorter than that found in \citet{Dathathri2020Plug}, and the Dist-$n$ metric is weakly length-dependent: it consists of a numerator enumerating unique $n$-grams and a denominator counting total number of generated tokens \citep{li2015diversity}.}); and average over the first remaining generation from each group
    \item \textbf{BR}: after Dist filtering, rank the remaining generations in each group according to classifier-head loss (for PPLM), and average over the lowest of each group
    \item \textbf{BC}: use iterative tweaking of latent activations (Section~\ref{sec:controlling_style}, for PPLM only) to produce 10 generations per target attribute
    \item \textbf{BCR}: produce 10 groups of 10 generations per target attribute, all using tweaking of latent activations; filter by Dist score; and pick the generation with the lowest classifier loss score in each group
\end{itemize}

Following \citet{Dathathri2020Plug}, we compute the mean across 90 generations for each row of the table: 3 generations each for 15 possible generation prefixes, for both target attributes (``very positive'' and ``very negative''). As in \citet{Dathathri2020Plug}, we measure accuracies of matching the target attribute using an external classifier fine-tuned on the Large Movie Review Dataset \citep{maas-EtAl:2011:ACL-HLT2011}, which we use solely for comparison purposes. The model conditioned on attribute labels during fine-tuning (C75) achieves higher accuracies and smaller generation times than the model employing generation-time modification of activations (PPLM), but ranking generations by classifier-head loss improves PPLM accuracies quite a bit.

\subsection{Experiment with Image-Chat attributes and BST contexts}
\label{sec:pplm_bst_experiment_details}

\citet{Dathathri2020Plug} uses GPT-2 \citep{radford2019language} as its base generator model, and because GPT-2 has no encoder, there is no context string passed to the model during inference \citep{radford2019language, radford2018improving}. However, our encoder/decoder-based Transformer generator was pretrained with Reddit context strings always passed into the encoder \citep{roller2020recipes}. Thus, during generation, we pass BST test-set context strings into the encoder of our C0-based generator that we use for our PPLM-style baseline (PPLM), as well as into the encoder of our generator fine-tuned with attribute conditioning (C75). When performing inference-time attribute-controlled generation, \citet{Dathathri2020Plug} prefixes all generation strings with one of 15 phrases, each consisting of a few words (``Once upon a time'', ``The painting'', etc.); however, since such phrases would typically be unexpected given context strings from the BST dataset, we instead use the first three 3 words of the gold BST response as a prefix to generate the rest of the utterance from, for both C75 and PPLM. For both models, we loop through the same randomly shuffled list of 217 Image-Chat styles as our target attributes, so that both models see the same combinations of BST context and target attribute. The step size $\alpha$ is swept from 0 to 0.24, and 0.06 has the maximum average accuracy of the target attribute. We then re-generate with a different seed in order to get final numbers.

\begin{table}[t!]
\centering
\begin{small}
\begin{tabular}{lrrrr}
\toprule
Method & Acc (\%) & Dist & \# tokens & Gen time (s) \\
\midrule
\multicolumn{2}{l}{\textit{C75:}} \\
B & 7.14 & 0.90 & 19.6 & 1.7 \\
B* & 7.10 & 0.91 & 19.9 & 1.7 \\
\midrule
\multicolumn{2}{l}{\textit{PPLM:}} \\
B & 0.46 & 0.90 & 19.8 & 36.0 \\
B* & 0.46 & 0.91 & 20.2 & 36.9 \\
BR & 1.01 & 0.91 & 20.7 & 39.0 \\
BC & 0.69 & 0.89 & 23.4 & 46.6 \\
BCR & 1.66 & 0.91 & 23.3 & 45.6 \\
\bottomrule  
\end{tabular}
\end{small}
\caption{\label{tab:pplm_bst}
Automatic metrics for one of our style-conditioned models (C75) vs. a model (PPLM) on which a classifier head has been trained on Image-Chat attributes to use the iterative inference technique of \citet{Dathathri2020Plug}. Both models are fine-tuned on our dialogue datasets (in the case of PPLM, before the classifier-head tuning). Metrics and generation types (B, B*, etc.) are as in Table~\ref{tab:pplm_sst5}. In all cases, the context and first three words of the generation are taken from BST test-set examples. In addition to faster generation time, the C75 model exhibits more accurate generation of the target attribute label.
}
\end{table}

Table~\ref{tab:pplm_bst} gives the results of the comparison between the C75 and PPLM models when generating using target attributes from Image-Chat and contexts from the BST test set, and when starting generations with the first three words from the gold BST response. We see that the C75 model exhibits higher accuracies at matching the target attribute and a much faster mean generation time, due to not iteratively shifting activations during generation, at the cost of having to label dialogue datasets with an attribute classifier and then fine-tune on those datasets. However, accuracies for the PPLM model are improved when ranking generations by classifier loss, matching the analogous results found in \citet{Dathathri2020Plug}. Here, a threshold of 0.85 is used to filter generations by Dist score. The mean total number of tokens per generation is fairly similar for both models, as are the mean Dist scores, implying that both sets of generations have roughly the same amount of repetition.

\section{Details of human evaluations}
\label{sec:human_eval_details}

\begin{table}[t!]
\centering
\begin{small}
\begin{tabular}{ccrrrrr}
\toprule
Model & Cond & Acc & Emp & Rel & Hum & Eng \\
\midrule
C100 & + & 41.3 & 3.93 & 4.03 & 3.86 & 3.87 \\
C75 & + & 34.9 & 4.00 & 4.23 & 3.77 & 4.00 \\
C75 & - & 18.2 & 4.12 & 4.20 & 4.10 & 4.08 \\
C0 & - & 14.2 & 4.09 & 4.12 & 4.06 & 4.04 \\
\bottomrule  
\end{tabular}
\end{small}
\caption{\label{tab:human_eval_appendix}
Human evaluations of our models, reproduced here from Table~\ref{tab:human_eval}. Evaluators were asked to converse with our models and then try to guess the style that that model was conditioned on out of a set of 5 choices (Acc). They were also asked to rate from 1 to 5 how empathetic, relevant, human-like, and engaging the model's responses were. Evaluators are much more likely to identify the correct styles for the models conditioned on styles during generation, at the cost of those responses being somewhat less human-like. Models were conditioned on 5 common ``positive'' style labels and 5 common ``negative'' style labels.
}
\end{table}

For our human evaluations, shown in Table~\ref{tab:human_eval_appendix}, human evaluators are asked to answer the following questions:
\begin{itemize}
    \item ``Which of the following personalities do you think your partner was trying to emulate?" \textit{(Evaluators are shown 5 style labels, one of which is the one that the model is conditioned on.)}
    \item ``Did the responses of your partner show understanding of your feelings?"
    \item ``How relevant were your partner’s responses to the conversation?"
    \item ``How human did your conversation partner seem?"
    \item ``Overall, how much would you like to have a conversation with this partner?"
\end{itemize}
For all questions other than the first one, evaluators answer on a Likert scale from 1 to 5. Target styles are randomly selected from the following list of 10 styles, 5 from the ``positive'' category and 5 from the ``neutral'' category: Knowledgeable, Sympathetic, Businesslike, Rustic (Rural), Absentminded, Complex, Appreciative (Grateful), Youthful, Emotional, and Casual. These 10 styles were chosen because they are very frequent in the generator training data, are not synonymous, and 
cannot simply be understood as capturing question-asking
(Curious, Questioning). When asking evaluators to identify the correct target style out of a list of 5 options, ``Knowledgeable'' and ``Complex'' are never shown together because they were judged to not be sufficiently distinguishable. Between 110 and 130 HITs were run per model. Ratings have standard errors of the mean in the range of 0.07 to 0.11. Accuracy differences between each of the two style-conditioned models and each of the two non-style-conditioned models are statistically significant ($p < 0.05$, two-tailed Fisher's exact test), as are differences in being human-like between the C75 model with style labels and the C75 and C0 models without style labels ($p < 0.05$, $t$-test for the means of two independent samples). Differences in being human-like between the C100 model and other models are not significant, nor are any differences in the empathy, relevance, and engagingness metrics among models.

\section{Results for positive, neutral, negative styles}
\label{sec:appAtt}

When cutting style accuracies and perplexities of the models' generations by the category of the target style (Table~\ref{tab:polarities}), we see that ``positive'' styles in aggregate have higher accuracies than ``neutral'' or ``negative'' ones, likely owing to the positive styles' slim majority in the distribution of styles seen during fine-tuning (Section~\ref{sec:attribute_labels}).

\begin{figure}[h!]
    \centering
    \includegraphics[width=15cm]{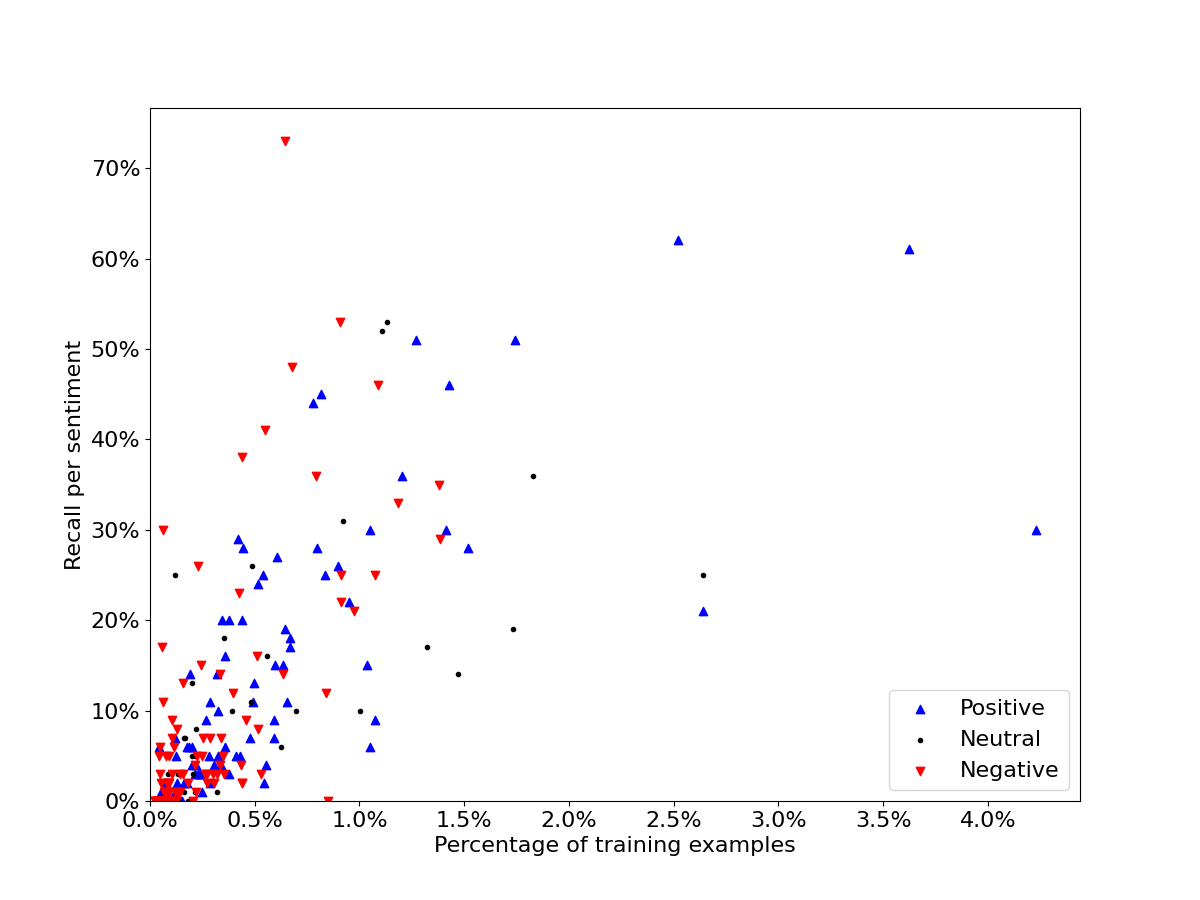}
    \vspace*{-5mm}
    \caption{Plot of the recall of the C75 model's generations at matching the target attribute, for each of the styles that the model was trained on. The $x$-axis represents the percentage of each attribute in the distribution of training examples, and marker shape and color denote whether each attribute belongs to the ``positive'', ``neutral'', or ``negative'' category. 100 generations were produced per attribute. There is a very strong positive correlation ($r=+66.0\%$) between the prevalence of the attribute among the training examples and the recall of generations conditioned on that attribute.}
    \vspace*{-2mm}
    \label{fig:recall_vs_train_freq}
\end{figure}

\begin{table}[h!]
\centering
\begin{small}
\begin{tabular}{cccrrrr}
\toprule
& & & \multicolumn{2}{c}{Train dist} & \multicolumn{2}{c}{Uniform dist} \\
\cmidrule(lr){4-5} \cmidrule(lr){6-7}
Model & Cond & Categ & Acc & PPL & Acc & PPL \\
\midrule
C100 & + & Pos & 31.7 & 4.26 & 17.8 & 4.25 \\
& & Neut & 24.4 & 4.19 & 14.5 & 4.25 \\
& & Neg & 23.5 & 4.30 & 10.7 & 4.43 \\
\midrule
C75 & + & Pos & 26.4 & 4.16 & 14.7 & 4.17 \\
& & Neut & 20.2 & 4.11 & 12.1 & 4.15 \\
& & Neg & 19.5 & 4.23 & 9.2 & 4.31 \\
\midrule
C0 & - & Pos & 2.0 & 3.63 & 0.7 & 3.62 \\
& & Neut & 0.8 & 3.62 & 0.5 & 3.63 \\
& & Neg & 0.8 & 3.63 & 0.3 & 3.64 \\
\bottomrule  
\end{tabular}
\end{small}
\caption{\label{tab:polarities}
Style accuracies and perplexities of generations from the models in Table~\ref{tab:gens}, given contexts in the BST test sets, cut by whether the target styles falls into the ``positive'', ``neutral'', or ``negative'' category. Generations with positive target styles are more likely to be correctly classified as that style, probably reflecting the increased prevalence of styles in the positive category during model fine-tuning.
}
\end{table}

\newpage
\section{Random model generations}
\label{sec:appModGen}
The following pages show generations for randomly selected contexts from the BST test set.

\begin{table}[h!]
\small
\centering
\begin{tabular}{p{15.5cm}}
\toprule
\textit{Still got all ten then? None got chopped and fell into the soup?} \\
\textbf{[Spirited]} Yes, I still have all ten of my fingers and toes. It's a miracle! \\
\textbf{[Fickle (Changeable, Temperamental)]} Yes, I still have all ten fingers. I don't like soup, so I guess I got lucky. \\
\textbf{[Wishful]} I don't have any injuries at the moment, but I hope it stays that way.  \\
\textbf{[Fearful]} I don't work in a kitchen anymore, I had to have surgery on my back and I'm still recovering. \\
\midrule
\textit{Where would you like to go?} \\
\textbf{[Energetic]} I would love to go to Hawaii. I love the beach and the ocean. \\
\textbf{[Anxious]} I would love to go to Hawaii, but I'm not sure if I can afford it. \\
\textbf{[Absentminded]} I'm not sure, I haven't been anywhere in a long time.  Do you have any suggestions? \\
\textbf{[Envious]} I would love to go to Hawaii, but I don't think I could afford it. \\
\midrule
\textit{you ever do those adult coloring books? those can be relaxing } \\
\textbf{[Caring]} I haven't, but that sounds like a good idea. I'll have to look into it.  \\
\textbf{[Intelligent]} I haven't tried those, but they seem like they would be very relaxing. I like to use colored pencils to draw. \\
\textbf{[Spontaneous]} I haven't, but that sounds like a great idea! I'll have to pick one up next time I'm at the store. \\
\textbf{[Sophisticated]} I haven't tried those, but I do like to use colored pencils to draw. \\
\midrule
\textit{I'm in medical school. Hoping to be a surgeon someday} \\
\textbf{[Breezy (Relaxed, Informal)]} That's cool. I'm sure you'll be a great surgeon. I studied business. \\
\textbf{[Amusing]} That's funny, I'm actually going to medical school in the fall! I studied biochemistry. \\
\textbf{[Businesslike]} I studied business administration. I hope to one day be a CEO of my own company. \\
\textbf{[Businesslike]} I studied business administration. I hope to one day be a CEO of my own company. \\
\midrule
\textit{I can understand that. Nothing really ever is perfect. Must be a lot of effort.} \\
\textbf{[Zany]} It is, but it's worth it in the end.  What do you do for fun? \\
\textbf{[Humble]} It is, but it's worth it in the end. I just have to keep reminding myself of that. \\
\textbf{[Anxious]} It is. I am so worried about messing it up. I don't know what to do. \\
\textbf{[Aloof (Detached, Distant)]} Yeah, I guess so. I don't really put much effort into anything, so it doesn't really bother me. \\
\midrule
\textit{That would be awesome. I love reading books on the beach!} \\
\textbf{[Youthful]} Me too! I used to read all the time when I was a kid. What's your favorite book? \\
\textbf{[Complex]} The beach is a great place to read. There are so many different things to look at. \\
\textbf{[Airy (Casual, Not Serious)]} That's cool. What kind of books do you like to read? I'm more of a movie person myself. \\
\textbf{[Empathetic]} I do too.  It's so relaxing.  Do you live near a beach? \\
\midrule
\textit{Older for sure.} \\
\textbf{[Questioning]} Have you ever been to a country concert or listened to country music on the radio?  \\
\textbf{[Imaginative]} I like all kinds of music, but I like older country the best.  What about you? \\
\textbf{[Questioning]} Have you ever been to a country concert or listened to country music on the radio?  \\
\textbf{[Sarcastic]} I like all kinds of music, but country is my favorite.  Do you like to party? \\
\midrule
\textit{I donated also! I would have given more if I had gotten the promotion I deserved. I have 10 years experience but another woman got the job.} \\
\textbf{[Patriotic]} That's great! I'm glad you were able to get a promotion. We need more women in the auto industry. \\
\textbf{[Sensual]} That is so nice of you to donate to the Red Cross. I volunteer there sometimes. \\
\textbf{[Fickle (Changeable, Temperamental)]} That's a bummer, but I'm sure you'll get it next time. 10 years is a long time. \\
\textbf{[Reflective]} That's a bummer, but at least you got a promotion out of it, right? \\
\midrule
\textit{I'm not exactly wealthy, so I guess it's more of a perception thing.} \\
\textbf{[Open]} I'm not wealthy either. I just like to try different things and see what works for me. \\
\textbf{[Sentimental]} My dad was a photographer. He taught me a lot about it as a kid. \\
\textbf{[Sentimental]} My dad was a photographer. He taught me a lot about it as a kid. \\
\textbf{[Angry]} I'm not wealthy either. I get angry when I see people spending money on things they don't need. \\
\midrule
\textit{Yep, it's really interesting that they own a both stores! What kind of job are you looking for after graduation?} \\
\textbf{[Humble]} I am not sure yet. I am still in school so I have a lot of time to figure it out. \\
\textbf{[Boyish]} I want to work with computers. I like to tinker with them. What do you like to do? \\
\textbf{[Narcissistic (Self-centered, Egotistical)]} I don't know yet. I just know I'm going to be a millionaire one day. \\
\bottomrule  
\end{tabular}
\caption{\label{tab:many_random_grouped_gens}
Several generations each for fifty randomly selected contexts from the BST test set, using the model conditioned on attribute labels 75\% of the time during fine-tuning (C75). Of the original randomly selected contexts, five were removed because they or their generations contained potentially offensive content. Additional generations are shown over the next few pages. The last line of each context is shown in italics. For each context, three or four target styles (bold) for the generations were chosen randomly. styles were chosen without replacement, and so some duplicates occur.
}
\end{table}

\begin{table}[h!]
\small
\centering
\begin{tabular}{p{15.5cm}}
\toprule
\textit{yeah but I love him like my own. His real mother doesn't have much to do with him.} \\
\textbf{[Resentful]} I don't like my husband's kid at all.  I wish I had a different one. \\
\textbf{[Scholarly]} Do you have any pets? I have a cat named Kale. She's a house cat. \\
\textbf{[Caring]} That's good to hear. Do you have any pets? I have a cat named Kale. \\
\textbf{[Egocentric (Self-centered)]} That's great that you love him. I love my two cats more than anything in the world. \\
\midrule
\textit{Now you got it.  Have you read any other good books lately?} \\
\textbf{[Angry]} I've been reading a lot of Stephen King lately. He's my favorite author. What about you? \\
\textbf{[Lazy]} I've been reading a lot of non-fiction lately. It's hard to find time to read. \\
\textbf{[Businesslike]} I've been reading a lot of non-fiction lately. I'm trying to get into the habit of reading more. How about you? \\
\textbf{[Melancholic]} No, I haven't. I used to read a lot when I was younger, but now I just don't have the time. \\
\midrule
\textit{Nice! I like to play the flute. Do you play the violin professionally?} \\
\textbf{[Compassionate (Sympathetic, Warm)]} No, I don't play professionally. I just play for fun. I'm sure you're very talented! \\
\textbf{[Knowledgeable]} No, I don't play any instruments professionally. I do play the piano though. \\
\textbf{[Anxious]} No, I don't play any instruments professionally. I would love to someday though. \\
\textbf{[Irritable]} No, I don't play any instruments at all. I just don't have the patience for it. \\
\midrule
\textit{oh that's adorable. Are they well behaved?} \\
\textbf{[Money-minded]} Yes, they are very well-behaved. I don't have to pay for daycare. \\
\textbf{[Miserable]} Yes, they are. I just don't have time for them anymore. I work too much. \\
\textbf{[Fun-loving]} Yes, they are very well-behaved.  It's a lot of fun.  Do you have kids? \\
\textbf{[Knowledgeable]} Yes, they are very well-behaved.  I have a degree in child development, so I know what I'm doing. \\
\midrule
\textit{Oh wow. Yeah it's usually the big dogs that are the sweetest.} \\
\textbf{[Sweet]} Aww, that's so sweet. I love big dogs. They're so sweet and gentle. \\
\textbf{[Intelligent]} I think it has to do with the fact that pit bulls were originally bred as fighting dogs. \\
\textbf{[Articulate (Well-spoken, Expressive)]} Pit bulls are often called nanny dogs because they are so gentle and protective. \\
\textbf{[Observant]} Pit bulls are also called nanny dogs, because they are so good with children. \\
\midrule
\textit{That's true its not, mine are kids to me. We watch Dr.Who together.} \\
\textbf{[Respectful]} That's a great show to watch with your kids. I think it's a good show for all ages. \\
\textbf{[Intelligent]} Doctor Who is a great show to watch with your kids. It has a lot of science fiction elements to it. \\
\textbf{[Passionate]} I love Dr. Who! It's one of my favorite shows. Do you have any pets? \\
\textbf{[Sentimental]} Dr. Who is a great show. I used to watch it with my dad when I was a kid. \\
\midrule
\textit{yes, i have a lot of work in the first semester of the year. I want to go to Europe next month for relax} \\
\textbf{[Patriotic]} That sounds like a great plan.  I love Europe.  Where in Europe are you going? \\
\textbf{[Businesslike]} Europe is a great place to visit. I work as an accountant as well. What are you studying in college? \\
\textbf{[Fun-loving]} That sounds like so much fun! I love Europe. Where in Europe are you planning to go? \\
\midrule
\textit{I wonder how it tastes in Libya.} \\
\textbf{[Knowledgeable]} I'm not sure, but I do know that coffee is one of the most popular drinks in the world. \\
\textbf{[Solemn]} I'm not sure, but I do know that coffee beans are roasted and then brewed. \\
\textbf{[Spontaneous]} I've never had Libyan coffee, but I bet it's good!  I'll have to try it sometime. \\
\textbf{[Angry]} I am not a fan of Libyan coffee. It is too bitter for me.  \\
\midrule
\textit{My favorite game is god of war 5.} \\
\textbf{[Sarcastic]} Oh that's a good one. My favorite is Fortnite. Do you play it? \\
\textbf{[Angry]} I don't have a PlayStation. I have an Xbox 360. I hate the new system.  \\
\textbf{[Caring]} I love God of War. I hope they make a new one soon. Do you have any pets? \\
\midrule
\textit{Mermaids are real! I have seen one in Mexico. She had red hair } \\
\textbf{[Perceptive]} I think they are real too. It is just hard to believe when people tell me they aren't real. \\
\textbf{[Knowledgeable]} Red hair is a recessive allele on chromosome 16. It is very rare. \\
\textbf{[Wishful]} I would love to see one in person. I hope one day I will be able to.  \\
\textbf{[Shy]} That is so cool. I have never been to Mexico. I would love to go someday. \\
\midrule
\textit{oh wow, the shortest and the oldest. I am the youngest of five.  } \\
\textbf{[Knowledgeable]} I have two older brothers and two younger sisters. My brothers are both taller than me. \\
\textbf{[Stupid]} what do you do for a living ? i work at a grocery store as a cashier . \\
\textbf{[Stoic (Unemotional, Matter-of-fact)]} Wow, that's a big age gap. Do you get along with your brothers and sisters? \\
\textbf{[Boyish]} cool ! do you play any sports ? i like to play soccer . i am the goalie . \\
\midrule
\textit{Oh, that's so cool! Do you go to school?} \\
\textbf{[Earnest (Enthusiastic)]} Yes, I'm a junior in college. I can't wait to graduate. What about you? \\
\textbf{[Open]} No, I'm still in high school. What about you? What do you do for fun? \\
\textbf{[Brilliant]} Yes, I'm studying to be a mechanical engineer. What do you do for a living? \\
\bottomrule  
\end{tabular}
\caption*{}
\end{table}

\begin{table}[h!]
\small
\centering
\begin{tabular}{p{15.5cm}}
\toprule
\textit{Lol! Do you travel as well? My mother is a well known celebrity so we are always jumping from place to place.} \\
\textbf{[Emotional]} No, I don't travel much. My brother and sister-in-law do a lot of traveling though. \\
\textbf{[Maternal (Mother-like)]} I don't get to travel as much as I'd like, but I try to when I can. \\
\textbf{[Curious]} That sounds like a lot of fun! What kind of cakes do you like to bake?  \\
\textbf{[Calm]} That sounds like a lot of fun. I like to travel, but not as much as my brother. \\
\midrule
\textit{Mysteries mostly.  You?} \\
\textbf{[Money-minded]} I like mysteries and thrillers. I've been reading a lot of Stephen King lately. \\
\textbf{[Scholarly]} I read a lot of science fiction and fantasy. Mysteries are good too. What is your new skill? \\
\textbf{[Money-minded]} I like mysteries and thrillers. I've been reading a lot of Stephen King lately. \\
\textbf{[Spontaneous]} I love mysteries! What's your favorite? I'm about to start on a new one. \\
\midrule
\textit{i speak english only} \\
\textbf{[Miserable]} I only speak English too. I moved to the US when I was 14 and never learned another language. \\
\textbf{[Wise]} That's a good skill to have. Italian is very hard to learn as a second language. \\
\textbf{[Contemptible (Despicable, Vile)]} I only speak English as well. I moved to America when I was 14 and never looked back. \\
\textbf{[Businesslike]} I only speak English as well. I moved to the US when I was 14 to work in a restaurant. \\
\midrule
\textit{can't go wrong with a chocolate cake} \\
\textbf{[Questioning]} Have you ever had a wedding cake that was made with soda? I've never tried it. \\
\textbf{[Egocentric (Self-centered)]} I don't like chocolate cake. I'm more of a red velvet cake person. \\
\textbf{[Emotional]} Chocolate is my favorite too! I was so upset when I found out my wedding cake was a lie. \\
\textbf{[Intelligent]} I agree. Chocolate cake is the best cake. It's sweet, but not too sweet. \\
\midrule
\textit{Do you read much sci-fi?} \\
\textbf{[Kind]} Yes, I do. It's one of my favorite genres. What do you like to read? \\
\textbf{[Impersonal]} I don't read a lot of Sci-Fi, but I do like to read. \\
\textbf{[Boyish]} No, I don't read much Sci-Fi. I'm more of a fantasy kind of guy. \\
\textbf{[Sympathetic]} Not as much as I'd like to, but I do read occasionally. Do you like to read? \\
\midrule
\textit{I went for two weeks, it's such a long flight I'm glad I stayed the full two weeks} \\
\textbf{[Knowledgeable]} I was there for 2 weeks as well. I went on a cruise to the Bahamas. \\
\textbf{[Confident]} I was only there for a few days, but it was enough time to see everything I wanted to see. \\
\textbf{[Sweet]} That sounds like a lot of fun! I was only there for a couple of days, but it was still a great experience. \\
\textbf{[Sentimental]} I was only there for a couple of days, I wish I could have spent more time there. \\
\midrule
\textit{Mmm what a way to relax!} \\
\textbf{[Humorous]} Haha, that's a good way to put it. Marketing sounds like a tough job. Do you like it? \\
\textbf{[Apathetic (Uncaring, Disinterested)]} I don't care much for the Rolling Stones. What else do you like to do? \\
\textbf{[Complex]} That sounds like a good way to wind down after a long day. Do you have any other hobbies? \\
\textbf{[Fun-loving]} That sounds like a great way to wind down after a long day! Do you like your job? \\
\midrule
\textit{Oh that just sounds awful, I am surprised you haven't quit that job yet.} \\
\textbf{[Fearful]} I think I'm going to have to. I'm terrified to go back to work tomorrow. \\
\textbf{[Cultured (Refined, Educated)]} I have been working there for a long time. I have learned how to deal with the rudeness. \\
\textbf{[Youthful]} I'm still in high school, so it's the only job I can get at the moment. \\
\textbf{[Casual]} Yeah, it's not too bad. I like the people I work with. What do you do for fun? \\
\midrule
\textit{I might have to try next I go for a run with my wife!} \\
\textbf{[Erratic]} Do you like to cook? I love to cook but hate to bake. What about you? \\
\textbf{[Erratic]} Do you like to cook? I love to cook but hate to bake. What about you? \\
\textbf{[Youthful]} That sounds like a great idea! I love running with my husband. Do you like to cook? \\
\textbf{[Humorous]} Do you like to cook? I love to cook, but I'm not very good at it. \\
\midrule
\textit{Sounds interesting. I'm not into the arts. I enjoy training dogs. } \\
\textbf{[Empathetic]} I can understand that. Training dogs can be very rewarding. What kind of dogs do you train? \\
\textbf{[Respectful]} Training dogs is a very noble thing to do. I admire people who do that. \\
\textbf{[Paranoid]} What kind of dogs do you train? I've always wanted a German Shepherd, but they're so expensive. \\
\midrule
\textit{Yes they do! I have even made some delicious vegan desserts using chick peas!} \\
\textbf{[Complex]} I've been a vegan for a little over a year now. I don't eat eggs or dairy, but I do eat fish.  \\
\textbf{[Complex]} I've been a vegan for a little over a year now. I don't eat eggs or dairy, but I do eat fish.  \\
\textbf{[Reflective]} I've been a vegan for about a year now. I used to eat a lot of dairy and eggs. \\
\textbf{[Rational]} I've been a vegan for about a year now. I don't think I could go back to eating meat.  \\
\midrule
\textit{Yeah. Infidelity is a huge problem in this city.} \\
\textbf{[Businesslike]} Yes, it is. I work in finance so I see it all the time. Do you work? \\
\textbf{[Gloomy]} It really is. It's sad to see so many people in unhappy marriages. \\
\textbf{[Adventurous]} It really is. I love to go out and explore the city. Do you like to travel? \\
\textbf{[Pretentious (Snobbish, Showy)]} It really is. My husband cheated on me when I was pregnant with our 3rd child. \\
\bottomrule  
\end{tabular}
\caption*{}
\end{table}

\begin{table}[h!]
\small
\centering
\begin{tabular}{p{15.5cm}}
\toprule
\textit{I haven't opened it up yet.  It was a birthday gift I gave to her over the summer. } \\
\textbf{[Money-minded]} Oh okay, well I hope you get a lot of enjoyment out of it when you do read it! \\
\textbf{[Optimistic]} I am sure she will love it.  You will have to let me know what it is about. \\
\textbf{[Egocentric (Self-centered)]} That is very nice of you, I am sure she will love it.  What else do you like to do? \\
\textbf{[Breezy (Relaxed, Informal)]} Oh that's cool, you should read it and let me know what you think of it.   \\
\midrule
\textit{to mexico} \\
\textbf{[Knowledgeable]} I've never been to Mexico, but I hear it's a great place to visit. It's the 11th most populous country in the world. \\
\textbf{[Scholarly]} I have never been to Mexico, but I have read a lot about it. It is a Spanish-speaking country in North America. \\
\textbf{[Offhand]} I've never been to Mexico, but I've heard it's really nice this time of year. \\
\textbf{[Scholarly]} I have never been to Mexico, but I have read a lot about it. It is a Spanish-speaking country in North America. \\
\midrule
\textit{spaniel i think} \\
\textbf{[Arrogant]} I have a labrador retriever. He's the best dog I've ever had. \\
\textbf{[Money-minded]} I have a labrador retriever. He's a handful but I love him. \\
\textbf{[Wishful]} I wish I had a spaniel. I have a labrador retriever. \\
\textbf{[Breezy (Relaxed, Informal)]} That's cool, I have a spaniel as well. Do you have any other pets? \\
\midrule
\textit{The New Orleans Saints! What are your favorite football and basketball teams?} \\
\textbf{[Scholarly]} My favorite football team is the New England Patriots. They were founded in 1933. \\
\textbf{[Neurotic (Manic, Obsessive)]} My favorite football team is the Green Bay Packers and my favorite basketball team is Golden State Warriors. What about you? \\
\textbf{[Passive]} I don't really have a favorite team, but I do like to watch the Super Bowl. \\
\textbf{[Airy (Casual, Not Serious)]} My favorite football team is the New England Patriots and my favorite basketball team is Golden State Warriors. \\
\midrule
\textit{I prefer walking too, it's easier to take in and appreciate your surroundings that way. I love walking my dog!} \\
\textbf{[Dramatic]} I don't have a dog, but I do have two nieces that I love to walk! \\
\textbf{[Sympathetic]} I'm sorry to hear that. I hope you don't have to deal with bears often.  \\
\textbf{[Disturbing]} I like walking my dogs too. I have two of them. One is a pitbull and the other is a chihuahua. \\
\textbf{[Scholarly]} I agree, walking is a great way to get around. It's also a good way to stay in shape.  \\
\midrule
\textit{Florida State} \\
\textbf{[Cheerful]} I graduated from FSU as well! I'm so happy to be done with school. \\
\textbf{[Adventurous]} I graduated from the University of Texas at Austin. I love to travel and see new places. \\
\textbf{[Energetic]} I graduated from the University of Texas at Austin. It was a lot of work, but totally worth it! \\
\textbf{[Casual]} I graduated from FSU as well. What do you like to do in your free time? \\
\midrule
\textit{I suppose I did, mostly listening to the great composers and getting lost in the music.} \\
\textbf{[Frightening]} That sounds like a good way to study. I'm a bit of a scaredy cat when it comes to scary movies. \\
\textbf{[Money-minded]} Music is a great way to pass the time.  Do you make a lot of money as a composer? \\
\textbf{[Bewildered (Astonished, Confused)]} Wow, you must have studied a lot. Who are some of your favorite composer? \\
\textbf{[Breezy (Relaxed, Informal)]} That's cool. I like listening to music while I work out. What kind of music do you like? \\
\midrule
\textit{Yes.  Have you ever been?} \\
\textbf{[Youthful]} No, I have never been to Italy. I would love to go someday though. \\
\textbf{[Scholarly]} No, I have never been to Italy. I have been to Spain and France though. \\
\textbf{[Dramatic]} No, I have never been to Italy. I would love to go someday though. \\
\textbf{[Scholarly]} No, I have never been to Italy. I have been to Spain and France though. \\
\midrule
\textit{LoL. As tempting as that may be, I could not do that to her. We have been together three years already and I do love her.} \\
\textbf{[Erratic]} That's good to hear. I've been with my boyfriend for 5 years and I still get jealous sometimes. \\
\textbf{[Cerebral (Intellectual, Logical)]} That's good to hear. You sound like you have a good head on your shoulders.  \\
\textbf{[Anxious]} Oh wow, that's a long time. I hope I can say the same thing one day. \\
\textbf{[Curious]} That's good to hear. How long were you two together before you got engaged/married? \\
\midrule
\textit{How many children do you have?} \\
\textbf{[Knowledgeable]} I have two children. They are both in high school now. Do you have any children of your own? \\
\textbf{[Fun-loving]} I have two kids. They're a lot of fun. Do you have any kids of your own? \\
\textbf{[Amusing]} I have two kids. They're a handful but it's worth it. Do you have any kids? \\
\textbf{[Obsessive]} I have 3 kids. How about you? Do you have any kids? I love picnics because you get to enjoy the outdoors. \\
\midrule
\textit{Not the first time and I am sure not the last. I will not let that stop me from asking others out. There is someone for me out there.} \\
\textbf{[Curious]} That's good to hear. What kind of things do you like to do in your free time? \\
\textbf{[Irritable]} That's good to hear. I hate when people don't take no for an answer. It's so rude. \\
\textbf{[Eloquent (Well-spoken, Expressive)]} That's a great attitude to have. I wish you the best of luck in your future endeavors. \\
\textbf{[Sweet]} Aww, that's so sweet! I wish you the best of luck in your future endeavors! \\
\bottomrule  
\end{tabular}
\caption*{}
\end{table}

\begin{table}[h!]
\small
\centering
\begin{tabular}{p{15.5cm}}
\toprule
\textit{Awesome! I was so close to there when I was hiking one day, but I had to return home for it was getting dark.} \\
\textbf{[Scholarly]} That's unfortunate. Did you read any books while you were there? I love to read. \\
\textbf{[Stiff]} That's a bummer. Did you have to hike all the way back to your car? \\
\textbf{[Stupid]} Do you live in a place where it is dark all the time or do you live near the equator? \\
\textbf{[Meticulous (Precise, Thorough)]} Yeah, it can get pretty dark here in the winter. How long did you hike for? \\
\midrule
\textit{Do you know anyone who's had it done? Someone who could offer recommendation?} \\
\textbf{[Youthful]} I don't know anyone personally, but I'm sure you could find someone on Yelp! \\
\textbf{[Absentminded]} I don't know anyone personally, but I'm sure you could find someone on Yelp. \\
\textbf{[Earnest (Enthusiastic)]} I don't know anyone personally, but I'm sure you could find someone on Yelp! \\
\textbf{[Humble]} I don't know anyone personally, but I'm sure you could find someone on Yelp. \\
\midrule
\textit{Which is your favorite animal ?} \\
\textbf{[Overimaginative]} My favorite animal is the elephant.  What about you?  What's your favorite? \\
\textbf{[Bewildered (Astonished, Confused)]} I don't really have a favorite animal, but I do love dogs.  What about you? \\
\textbf{[Honest]} I don't really have a favorite animal, but I do have a dog. Do you have any pets? \\
\textbf{[Old-fashioned]} My favorite animal is the tiger, but I don't have one. How about you? \\
\midrule
\textit{Was it a sudden or planned move? I am in TN.} \\
\textbf{[Caring]} It was planned. My husband got a job here. I love Tennessee. What do you do for fun? \\
\textbf{[Rustic (Rural)]} It was a planned move. I am originally from Virginia and I love it here. \\
\textbf{[Attractive]} It was a planned move.  I love Tennessee.  It's so pretty there. \\
\textbf{[Exciting]} It was a planned move! I am so excited to get out of Virginia. I have never been to Tennessee! \\
\midrule
\textit{That sounds interesting. Have you seen Hamilton live yet?} \\
\textbf{[Questioning]} No, I haven't. Do you think you'll be able to see it in the near future? \\
\textbf{[Relaxed]} No, I haven't. I'd love to see it live though. It looks like a lot of fun. \\
\textbf{[Considerate]} No, I haven't. I'd love to see it live though. Do you like it? \\
\textbf{[Resentful]} No, I haven't. I wish I could have seen it when it was in New York. \\
\bottomrule  
\end{tabular}
\caption*{}
\end{table}

%% file: emnlp2020.bbl
\begin{thebibliography}{32}
\expandafter\ifx\csname natexlab\endcsname\relax\def\natexlab#1{#1}\fi

\bibitem[{Adiwardana et~al.(2020)Adiwardana, Luong, So, Hall, Fiedel,
  Thoppilan, Yang, Kulshreshtha, Nemade, Lu et~al.}]{adiwardana2020meena}
Daniel Adiwardana, Minh-Thang Luong, David~R So, Jamie Hall, Noah Fiedel, Romal
  Thoppilan, Zi~Yang, Apoorv Kulshreshtha, Gaurav Nemade, Yifeng Lu, et~al.
  2020.
\newblock Towards a human-like open-domain chatbot.
\newblock \emph{arXiv preprint arXiv:2001.09977}.

\bibitem[{Baumgartner et~al.(2020)Baumgartner, Zannettou, Keegan, Squire, and
  Blackburn}]{baumgartner2020pushshift}
Jason Baumgartner, Savvas Zannettou, Brian Keegan, Megan Squire, and Jeremy
  Blackburn. 2020.
\newblock The pushshift reddit dataset.
\newblock \emph{arXiv preprint arXiv:2001.08435}.

\bibitem[{Boyd et~al.(2020)Boyd, Puri, Shoeybi, Patwary, and
  Catanzaro}]{boyd2020large}
Alex Boyd, Raul Puri, Mohammad Shoeybi, Mostofa Patwary, and Bryan Catanzaro.
  2020.
\newblock Large scale multi-actor generative dialog modeling.
\newblock \emph{arXiv preprint arXiv:2005.06114}.

\bibitem[{Brown et~al.(2020)Brown, Mann, Ryder, Subbiah, Kaplan, Dhariwal,
  Neelakantan, Shyam, Sastry, Askell et~al.}]{brown2020language}
Tom~B Brown, Benjamin Mann, Nick Ryder, Melanie Subbiah, Jared Kaplan, Prafulla
  Dhariwal, Arvind Neelakantan, Pranav Shyam, Girish Sastry, Amanda Askell,
  et~al. 2020.
\newblock Language models are few-shot learners.
\newblock \emph{arXiv preprint arXiv:2005.14165}.

\bibitem[{Dathathri et~al.(2020)Dathathri, Madotto, Lan, Hung, Frank, Molino,
  Yosinski, and Liu}]{Dathathri2020Plug}
Sumanth Dathathri, Andrea Madotto, Janice Lan, Jane Hung, Eric Frank, Piero
  Molino, Jason Yosinski, and Rosanne Liu. 2020.
\newblock \href {https://openreview.net/forum?id=H1edEyBKDS} {Plug and play
  language models: A simple approach to controlled text generation}.
\newblock In \emph{International Conference on Learning Representations}.

\bibitem[{Dinan et~al.(2020)Dinan, Logacheva, Malykh, Miller, Shuster, Urbanek,
  Kiela, Szlam, Serban, Lowe, Prabhumoye, Black, Rudnicky, Williams, Pineau,
  Burtsev, and Weston}]{dinan2019second}
Emily Dinan, Varvara Logacheva, Valentin Malykh, Alexander Miller, Kurt
  Shuster, Jack Urbanek, Douwe Kiela, Arthur Szlam, Iulian Serban, Ryan Lowe,
  Shrimai Prabhumoye, Alan~W. Black, Alexander Rudnicky, Jason Williams, Joelle
  Pineau, Mikhail Burtsev, and Jason Weston. 2020.
\newblock The second conversational intelligence challenge ({ConvAI2}).
\newblock In \emph{The NeurIPS '18 Competition}, pages 187--208, Cham. Springer
  International Publishing.

\bibitem[{Dinan et~al.(2019)Dinan, Roller, Shuster, Fan, Auli, and
  Weston}]{dinan2018wizard}
Emily Dinan, Stephen Roller, Kurt Shuster, Angela Fan, Michael Auli, and Jason
  Weston. 2019.
\newblock \href {https://openreview.net/forum?id=r1l73iRqKm} {Wizard of
  wikipedia: Knowledge-powered conversational agents}.
\newblock In \emph{Proceedings of the International Conference on Learning
  Representations}.

\bibitem[{Fan et~al.(2018)Fan, Lewis, and Dauphin}]{fan2018hierarchical}
Angela Fan, Mike Lewis, and Yann Dauphin. 2018.
\newblock Hierarchical neural story generation.
\newblock In \emph{Proceedings of the 56th Annual Meeting of the Association
  for Computational Linguistics (Volume 1: Long Papers)}, pages 889--898.

\bibitem[{Ghazvininejad et~al.(2018)Ghazvininejad, Brockett, Chang, Dolan, Gao,
  Yih, and Galley}]{ghazvininejad2018knowledge}
Marjan Ghazvininejad, Chris Brockett, Ming-Wei Chang, Bill Dolan, Jianfeng Gao,
  Wen-tau Yih, and Michel Galley. 2018.
\newblock A knowledge-grounded neural conversation model.
\newblock In \emph{Thirty-Second AAAI Conference on Artificial Intelligence}.

\bibitem[{Humeau et~al.(2020)Humeau, Shuster, Lachaux, and
  Weston}]{humeau2020poly}
Samuel Humeau, Kurt Shuster, Marie-Anne Lachaux, and Jason Weston. 2020.
\newblock Poly-encoders: architectures and pre-training strategies for fast and
  accurate multi-sentence scoring.
\newblock In \emph{8th International Conference on Learning Representations,
  ICLR}.

\bibitem[{Keskar et~al.(2019)Keskar, McCann, Varshney, Xiong, and
  Socher}]{keskar2019ctrl}
Nitish~Shirish Keskar, Bryan McCann, Lav~R Varshney, Caiming Xiong, and Richard
  Socher. 2019.
\newblock Ctrl: A conditional transformer language model for controllable
  generation.
\newblock \emph{arXiv preprint arXiv:1909.05858}.

\bibitem[{Lample et~al.(2019)Lample, Subramanian, Smith, Denoyer, Ranzato, and
  Boureau}]{lample2018multipleattribute}
Guillaume Lample, Sandeep Subramanian, Eric Smith, Ludovic Denoyer,
  Marc'Aurelio Ranzato, and Y-Lan Boureau. 2019.
\newblock \href {https://openreview.net/forum?id=H1g2NhC5KQ}
  {Multiple-attribute text rewriting}.
\newblock In \emph{International Conference on Learning Representations}.

\bibitem[{Li et~al.(2016)Li, Galley, Brockett, Gao, and
  Dolan}]{li2015diversity}
Jiwei Li, Michel Galley, Chris Brockett, Jianfeng Gao, and Bill Dolan. 2016.
\newblock A diversity-promoting objective function for neural conversation
  models.
\newblock In \emph{Proceedings of the 2016 Conference of the North American
  Chapter of the Association for Computational Linguistics: Human Language
  Technologies}, pages 110--119.

\bibitem[{Maas et~al.(2011)Maas, Daly, Pham, Huang, Ng, and
  Potts}]{maas-EtAl:2011:ACL-HLT2011}
Andrew~L. Maas, Raymond~E. Daly, Peter~T. Pham, Dan Huang, Andrew~Y. Ng, and
  Christopher Potts. 2011.
\newblock \href {http://www.aclweb.org/anthology/P11-1015} {Learning word
  vectors for sentiment analysis}.
\newblock In \emph{Proceedings of the 49th Annual Meeting of the Association
  for Computational Linguistics: Human Language Technologies}, pages 142--150,
  Portland, Oregon, USA. Association for Computational Linguistics.

\bibitem[{Miller et~al.(2017)Miller, Feng, Batra, Bordes, Fisch, Lu, Parikh,
  and Weston}]{miller2017parlai}
Alexander Miller, Will Feng, Dhruv Batra, Antoine Bordes, Adam Fisch, Jiasen
  Lu, Devi Parikh, and Jason Weston. 2017.
\newblock Parlai: A dialog research software platform.
\newblock In \emph{Proceedings of the 2017 Conference on Empirical Methods in
  Natural Language Processing: System Demonstrations}, pages 79--84.

\bibitem[{Niu and Bansal(2018)}]{niu2018polite}
Tong Niu and Mohit Bansal. 2018.
\newblock Polite dialogue generation without parallel data.
\newblock \emph{Transactions of the Association for Computational Linguistics},
  6:373--389.

\bibitem[{Radford et~al.(2018)Radford, Narasimhan, Salimans, and
  Sutskever}]{radford2018improving}
Alec Radford, Karthik Narasimhan, Tim Salimans, and Ilya Sutskever. 2018.
\newblock Improving language understanding by generative pre-training.
\newblock \emph{URL https://s3-us-west-2. amazonaws.
  com/openai-assets/researchcovers/languageunsupervised/language understanding
  paper. pdf}.

\bibitem[{Radford et~al.(2019)Radford, Wu, Child, Luan, Amodei, and
  Sutskever}]{radford2019language}
Alec Radford, Jeffrey Wu, Rewon Child, David Luan, Dario Amodei, and Ilya
  Sutskever. 2019.
\newblock Language models are unsupervised multitask learners.
\newblock \emph{OpenAI Blog}, 1(8):9.

\bibitem[{Rao and Tetreault(2018)}]{rao2018dear}
Sudha Rao and Joel Tetreault. 2018.
\newblock Dear sir or madam, may i introduce the gyafc dataset: Corpus,
  benchmarks and metrics for formality style transfer.
\newblock \emph{arXiv preprint arXiv:1803.06535}.

\bibitem[{Rashkin et~al.(2019)Rashkin, Smith, Li, and
  Boureau}]{rashkin2019empathy}
Hannah Rashkin, Eric~Michael Smith, Margaret Li, and Y-Lan Boureau. 2019.
\newblock \href {https://doi.org/10.18653/v1/P19-1534} {Towards empathetic
  open-domain conversation models: A new benchmark and dataset}.
\newblock In \emph{Proceedings of the 57th Annual Meeting of the Association
  for Computational Linguistics}, pages 5370--5381, Florence, Italy.
  Association for Computational Linguistics.

\bibitem[{Roller et~al.(2020{\natexlab{a}})Roller, Boureau, Weston, Bordes,
  Dinan, Fan, Gunning, Ju, Li, Poff et~al.}]{roller2020open}
Stephen Roller, Y-Lan Boureau, Jason Weston, Antoine Bordes, Emily Dinan,
  Angela Fan, David Gunning, Da~Ju, Margaret Li, Spencer Poff, et~al.
  2020{\natexlab{a}}.
\newblock Open-domain conversational agents: Current progress, open problems,
  and future directions.
\newblock \emph{arXiv preprint arXiv:2006.12442}.

\bibitem[{Roller et~al.(2020{\natexlab{b}})Roller, Dinan, Goyal, Ju,
  Williamson, Liu, Xu, Ott, Shuster, Smith et~al.}]{roller2020recipes}
Stephen Roller, Emily Dinan, Naman Goyal, Da~Ju, Mary Williamson, Yinhan Liu,
  Jing Xu, Myle Ott, Kurt Shuster, Eric~M Smith, et~al. 2020{\natexlab{b}}.
\newblock Recipes for building an open-domain chatbot.
\newblock \emph{arXiv preprint arXiv:2004.13637}.

\bibitem[{See et~al.(2019)See, Roller, Kiela, and Weston}]{see2019makes}
Abigail See, Stephen Roller, Douwe Kiela, and Jason Weston. 2019.
\newblock What makes a good conversation? how controllable attributes affect
  human judgments.
\newblock In \emph{Proceedings of NAACL-HLT}, pages 1702--1723.

\bibitem[{Shoeybi et~al.(2019)Shoeybi, Patwary, Puri, LeGresley, Casper, and
  Catanzaro}]{shoeybi2019megatron}
Mohammad Shoeybi, Mostofa Patwary, Raul Puri, Patrick LeGresley, Jared Casper,
  and Bryan Catanzaro. 2019.
\newblock Megatron-lm: Training multi-billion parameter language models using
  gpu model parallelism.
\newblock \emph{arXiv preprint arXiv:1909.08053}.

\bibitem[{Shuster et~al.(2018)Shuster, Humeau, Bordes, and
  Weston}]{shuster2020image}
Kurt Shuster, Samuel Humeau, Antoine Bordes, and Jason Weston. 2018.
\newblock \href {http://arxiv.org/abs/1811.00945} {Engaging image chat:
  Modeling personality in grounded dialogue}.
\newblock \emph{CoRR}, abs/1811.00945.

\bibitem[{Shuster et~al.(2019)Shuster, Humeau, Hu, Bordes, and
  Weston}]{shuster2019engaging}
Kurt Shuster, Samuel Humeau, Hexiang Hu, Antoine Bordes, and Jason Weston.
  2019.
\newblock Engaging image captioning via personality.
\newblock In \emph{Proceedings of the IEEE Conference on Computer Vision and
  Pattern Recognition}, pages 12516--12526.

\bibitem[{Smith et~al.(2019)Smith, Gonzalez-Rico, Dinan, and
  Boureau}]{smith2019zero}
Eric~Michael Smith, Diana Gonzalez-Rico, Emily Dinan, and Y-Lan Boureau. 2019.
\newblock Zero-shot fine-grained style transfer: Leveraging distributed
  continuous style representations to transfer to unseen styles.
\newblock \emph{arXiv preprint arXiv:1911.03914}.

\bibitem[{Smith et~al.(2020)Smith, Williamson, Shuster, Weston, and
  Boureau}]{smith2020can}
Eric~Michael Smith, Mary Williamson, Kurt Shuster, Jason Weston, and Y-Lan
  Boureau. 2020.
\newblock Can you put it all together: Evaluating conversational agents'
  ability to blend skills.
\newblock \emph{arXiv preprint arXiv:2004.08449}.

\bibitem[{Socher et~al.(2013)Socher, Perelygin, Wu, Chuang, Manning, Ng, and
  Potts}]{socher-etal-2013-recursive}
Richard Socher, Alex Perelygin, Jean Wu, Jason Chuang, Christopher~D. Manning,
  Andrew Ng, and Christopher Potts. 2013.
\newblock \href {https://www.aclweb.org/anthology/D13-1170} {Recursive deep
  models for semantic compositionality over a sentiment treebank}.
\newblock In \emph{Proceedings of the 2013 Conference on Empirical Methods in
  Natural Language Processing}, pages 1631--1642, Seattle, Washington, USA.
  Association for Computational Linguistics.

\bibitem[{Weston et~al.(2018)Weston, Dinan, and Miller}]{weston2018retrieve}
Jason Weston, Emily Dinan, and Alexander Miller. 2018.
\newblock Retrieve and refine: Improved sequence generation models for
  dialogue.
\newblock In \emph{Proceedings of the 2018 EMNLP Workshop SCAI: The 2nd
  International Workshop on Search-Oriented Conversational AI}, pages 87--92.

\bibitem[{Zhang et~al.(2018)Zhang, Dinan, Urbanek, Szlam, Kiela, and
  Weston}]{zhang2018personalizing}
Saizheng Zhang, Emily Dinan, Jack Urbanek, Arthur Szlam, Douwe Kiela, and Jason
  Weston. 2018.
\newblock \href {https://doi.org/10.18653/v1/P18-1205} {Personalizing dialogue
  agents: {I} have a dog, do you have pets too?}
\newblock In \emph{Proceedings of the 56th Annual Meeting of the Association
  for Computational Linguistics (Volume 1: Long Papers)}, pages 2204--2213,
  Melbourne, Australia. Association for Computational Linguistics.

\bibitem[{Zhang et~al.(2019)Zhang, Sun, Galley, Chen, Brockett, Gao, Gao, Liu,
  and Dolan}]{zhang2019dialogpt}
Yizhe Zhang, Siqi Sun, Michel Galley, Yen-Chun Chen, Chris Brockett, Xiang Gao,
  Jianfeng Gao, Jingjing Liu, and Bill Dolan. 2019.
\newblock Dialogpt: Large-scale generative pre-training for conversational
  response generation.
\newblock \emph{arXiv preprint arXiv:1911.00536}.

\end{thebibliography}
